%% file: LANet.tex
\documentclass{egpubl}
\usepackage{pg2021}
\SpecialIssuePaper         
\usepackage[T1]{fontenc}
\usepackage{dfadobe}  
\usepackage{cite}  
\BibtexOrBiblatex
\electronicVersion
\PrintedOrElectronic
\ifpdf \usepackage[pdftex]{graphicx} \pdfcompresslevel=9
\else \usepackage[dvips]{graphicx} \fi
\usepackage{egweblnk}

\usepackage{graphicx}
\usepackage{amsmath}
\usepackage{multirow}

\usepackage{color}

\usepackage[misc]{ifsym}
\usepackage[toc,page]{appendix}

\title[Luminance Attentive Networks for HDR Image and Panorama Reconstruction]%
      {Luminance Attentive Networks for HDR Image and \\
      Panorama Reconstruction}

\author[H. Yu \& W. Liu \& C. Long \& B. Dong \& Q. Zou \& C. Xiao]
{\parbox{0.98\textwidth}{\centering 
Hanning Yu$^{1*}$\orcid{0000-0002-6619-4982},
Wentao Liu$^{1*}$\orcid{0000-0002-0584-9113},
Chengjiang Long$^{2}$$^{\small\dag}$\orcid{0000-0003-1584-7290},
Bo Dong$^{3}$\orcid{0000-0001-9189-9506}, 
Qin Zou$^{1}$\orcid{0000-0001-7955-0782},
Chunxia Xiao$^{1}$\thanks{This work was co-supervised by Chengjiang Long and Chunxia Xiao. \\
$*$ \text{ }Hanning Yu and Wentao Liu are joint first authors.\\
$\small\ddag$ \text{ }Chunxia Xiao is the corresponding author.
}$^{\small\ddag}$\orcid{0000-0002-4526-6297}
        }
        \\
{\parbox{0.98\textwidth}{\centering $^1$
School of Computer Science, Wuhan University, Wuhan, Hubei, China 430072\\
         $^2$JD Finance America Corporation, Mountain View, CA, USA 94043\\
         $^3$SRI international, Princeton, NJ, USA 08540 \\
    fishaning@whu.edu.cn, liu136583410@gmail.com, cjfykx@gmail.com, dongshuhao12@gmail.com, qzou@whu.edu.cn, cxxiao@whu.edu.cn
       }
}
}

\begin{document}
\maketitle

\begin{abstract}
  It is very challenging to reconstruct a high dynamic range (HDR) from a low dynamic range (LDR) image as an ill-posed problem. This paper proposes a luminance attentive network named LANet for HDR reconstruction from a single LDR image. Our method is based on two fundamental observations: (1) HDR images stored in relative luminance are scale-invariant, which means the HDR images will hold the same information when multiplied by any positive real number. Based on this observation, we propose a novel normalization method called " {\em HDR calibration} " for HDR images stored in relative luminance, calibrating HDR images into a similar luminance scale according to the LDR images. (2) The main difference between HDR images and LDR images is in under-/over-exposed areas, especially those highlighted. Following this observation, we propose a luminance attention module with a two-stream structure for LANet to pay more attention to the under-/over-exposed areas. In addition, we propose an extended network called panoLANet for HDR panorama reconstruction from an LDR panorama and build a dual-net structure for panoLANet to solve the distortion problem caused by the equirectangular panorama. Extensive experiments show that our proposed approach LANet can reconstruct visually convincing HDR images and demonstrate its superiority over state-of-the-art approaches in terms of all metrics in inverse tone mapping. The image-based lighting application with our proposed panoLANet also demonstrates that our method can simulate natural scene lighting using only LDR panorama. Our source code is available at {\color{magenta}{\url{https://github.com/LWT3437/LANet}}}.



\end{abstract}

\section{Introduction}

With a limited dynamic range, a low dynamic range (LDR) image captured by a standard digital camera can not represent the real luminance of the scene and suffers under-/over-exposure problems. High dynamic range (HDR) images solve this problem by recording HDR information of the scene. As HDR images can provide larger luminance variance and contain richer details than LDR images, people can benefit significantly from HDR images in visual perception. HDR images have also been widely used in image-based lighting (IBL) technology to provide more realistic rendering results. A common way to construct an HDR image is by merging a stack of bracketed exposure LDR images. However, dynamic scenes require significant efforts to produce equal-quality results like static scenes, as the moving parts require special treatments. Therefore, generating an HDR image from a single LDR image is getting more attention. However, reconstructing a high-quality HDR image from a single LDR image is also very challenging.

Most previous methods leverage inverse tone-mapping to solve the problem through traditional image processing technology~\cite{akyuz2007hdr,kovaleski2014high,masia2009evaluation,masia2017dynamic,rempel2007ldr2hdr}. These methods exploit individual heuristics or manual intervention to enhance LDR images, which cannot sufficiently compensate for lost data caused by color quantization and under-/over-exposure. Recently, deep learning-based approaches ~\cite{eilertsen2017hdr,endo2017deep,kuo2012content,marnerides2018expandnet,lee2018deep,wang2019deep} are proposed to automatically infer a statistically plausible HDR image from a single input LDR image. Although these deep learning-based approaches can bring more appealing results, they still have room to be improved. More importantly, most existing HDR datasets are in the relative luminance domain instead of the absolute one, bringing scale ambiguity to the training process. Previous methods used the general maximum normalization method to process HDR images, but this did not solve the scale ambiguity of luminance. This motivates us to investigate how to remove the uncertainty caused by the different luminance scales and achieve better reconstruction results.

\begin{figure}[tbp]
  \centering
  \mbox{} 
  \includegraphics[width=0.233\textwidth]{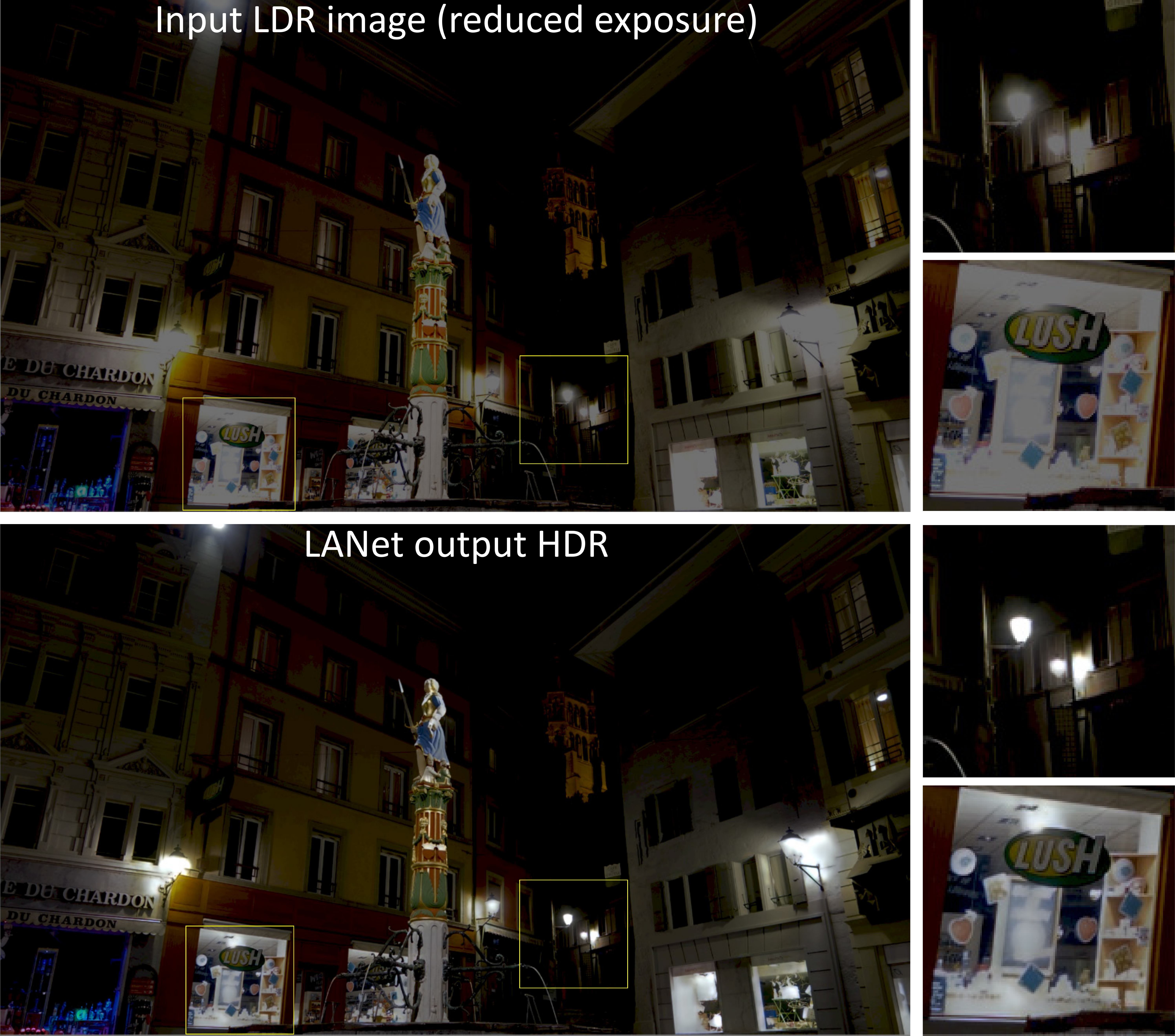}
  \vspace{-0.3cm}
  \hfill
  \includegraphics[width=0.233\textwidth]{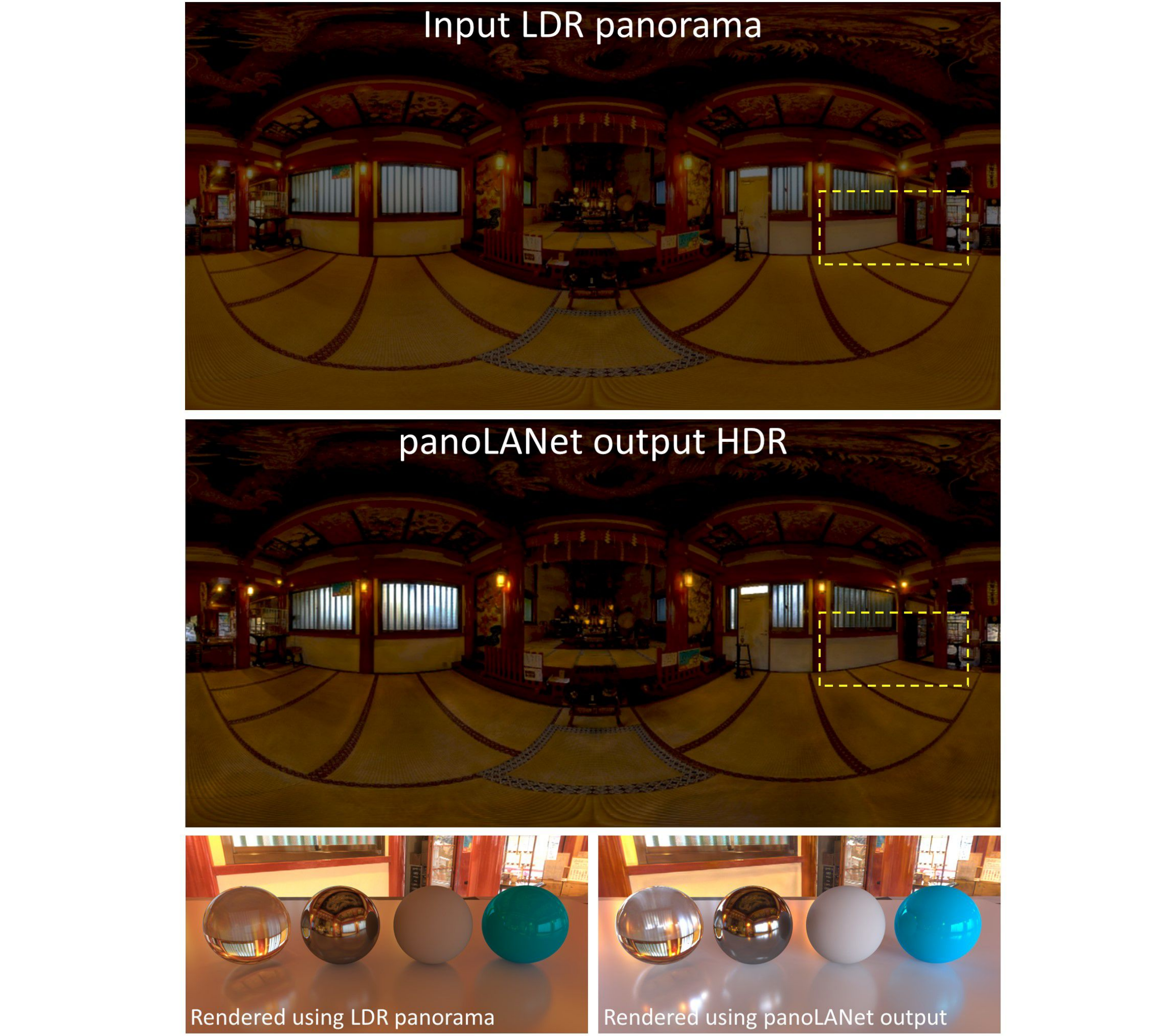}
  \mbox{}\\
  \hspace{0.005\textwidth}\mbox{(a) Inverse Tone Mapping} \hspace{0.06\textwidth}      \mbox{(b) Application on IBL}
  
    \caption{Example results of our proposed method. The left is the reconstruction result of our proposed LANet on a single image. The right results are rendered by original LDR panorama and reconstructed HDR panorama using our panoLANet, respectively. Note that the images of subfigure (a) are mapped using LuminanceHDR software in the same dynamic range and we reduced the exposure of original result for visualizing.} 
\label{represents}
\vspace{-0.5cm}
\end{figure}

We observe that two different HDR images may store the same scene, as they are saved in different luminance scales. That is to say, the difference between two HDR images is scale-invariant in the relative luminance domain, which means multiplying an HDR image by any positive real number will not change the information it represents. In such cases, to deal with the scale ambiguity problem, we propose an HDR normalizing approach to calibrate all HDR images into a similar luminance scale and introduce the scale-invariant loss~\cite{eigen2014depth} into the HDR learning task. 

We also observe that the main error between LDR images and HDR images occurs in the under-/over-exposed areas, especially in over-exposed areas. This requires the network to reconstruct the high dynamic range based on the image's brightness. Here we introduce the attention mechanism with two different network structures to achieve local spatial attention for general images and panoramas respectively, enabling the networks to focus more on over-exposed areas while avoiding errors to other areas as much as possible.

\begin{figure*}[ht!]
  \centering
  \includegraphics[width=0.95\textwidth]{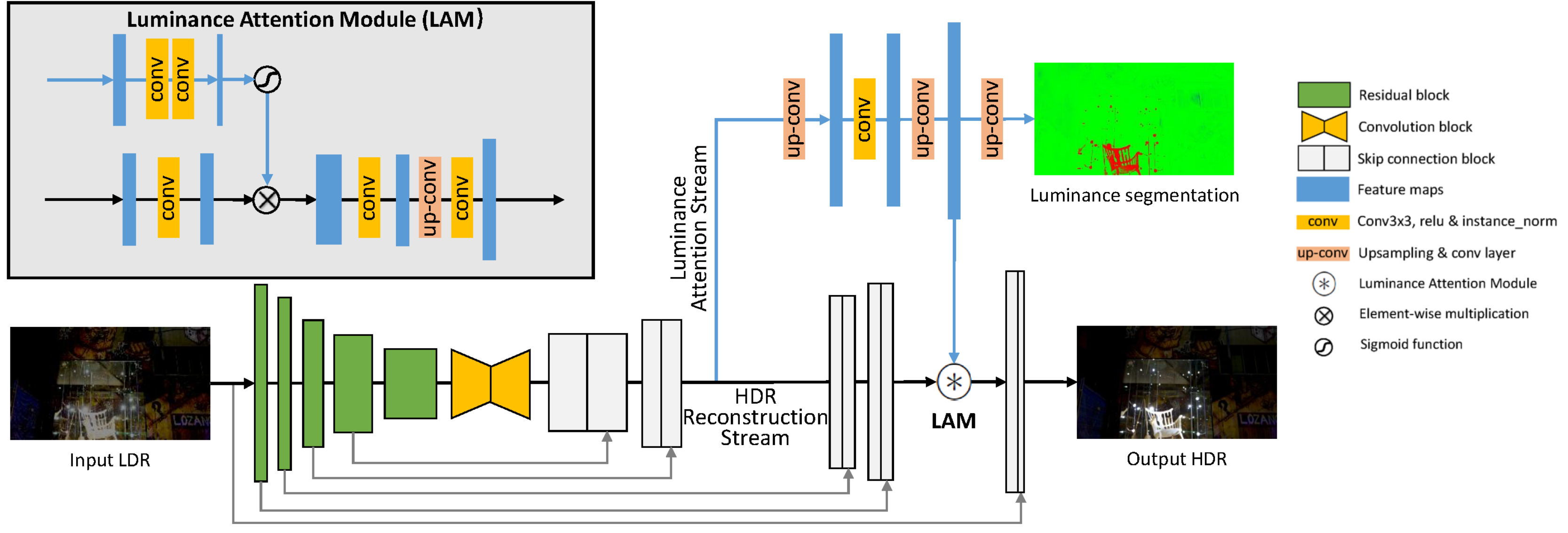}
  \caption{The overview of our LANet network. The convolution block consists of two convolutional layers with stride 2 and two up-conv layers. The skip-connection block first upsamples the previous layer's output and through two $3\times3$ convolution layers to keep the number of channels consistent with the jump layer. Meanwhile, a $3\times3$ convolution layer is applied to the output of the skip layer. Finally, the two convolutional results are concatenated and through an $1\times1$ convolution layer to obtain the output result.}
  \label{architecture-1}
  \vspace{-0.4cm}
\end{figure*}

In this paper, based on the above two observations, we propose Luminance Attentive Network (LANet) to construct an HDR image from a single LDR image, as shown in Fig.~\ref{architecture-1}. The LANet is designed as a multi-task network with two streams, named {\em luminance attention stream} and {\em HDR reconstruction stream}. The luminance attention stream is designed for network to learn to obtain a spatial weighted attention map about the luminance distribution. This design exploits estimated luminance segmentation as an auxiliary task to supervise the attention weights, and a novel luminance attention module is proposed to guide the reconstruction process paying more attention to those under-/over-exposed areas. 

To validate the potential of our proposed LANet, we further propose its extension network called panoLANet (see Fig.~\ref{architecture-2}) to reconstruct HDR panoramas for IBL rendering. The panoLANet consists of a {\em ceiling luminance branch} and a {\em panorama reconstruction branch}, in which the ceiling luminance branch is used to reconstruct the highlighted area in the upper half of the scene. We rotate and apply a perspective transformation on the panorama to produce a ceiling-view image to solve the distortion problem of the equirectangular panorama and use skip-connection with gated attention to sharing the information between the two branches.

It is worth mentioning here that although the usage of HDR panoramas is different from general HDR images, we still design these two networks, LANet and panoLANet, with similar structures and design principles. Both of these two networks can use the same data processing and training methods. LANet, as a general inverse tone mapping method, can achieve the HDR reconstruction task from the most common images. Meanwhile, as an extension of LANet, panoLANet can be applied to IBL rendering technology. As two examples, Fig.~\ref{represents} (a) shows our proposed LANet can well handle the areas under-/over-exposed areas, and Fig.~\ref{represents} (b) shows that our proposed panoLANet can generate high-quality results for applications on image-based lighting.

To sum up, the main contributions of this paper are three-fold: 
\begin{itemize}
  \item[$\bullet$] We propose an end-to-end trainable luminance attentive network called LANet with two streams for HDR reconstruction and performs better quantitatively and qualitatively than prior work.
  \item[$\bullet$] We propose a novel HDR calibration method, calibrating HDR images into a similar luminance scale according to the LDR images.
  \item[$\bullet$] We extend LANet to the panoLANet for HDR reconstruction from an LDR panorama.
\end{itemize}

We conduct experiments on public datasets~\cite{nemoto2015visual,hdrlabs} with both real LDR inputs and ground truth HDR references. Extensive experimental results have powerfully demonstrated the superiority of our proposed approaches LANet and panoLANet over state-of-the-art approaches.

\section{Related Work}

The related work can be divided into three categories, {\em i.e.}, {\em custom-made hardware}, {\em inverse Tone-Mapping Operators (iTMO)}, and {\em Deep learning-based approaches}.

{\textbf{Custom-made hardware}} is one way to generate high-quality HDR images from a single shot through specially designed hardware ({\em e.g.}, coded aperture and beam-splitter)~\cite{Tocci2011AVH,Mcguire2007,10.1145/2661229.2661260,HKU15}. However, custom-made hardware is more expensive and high-profile, making it hard to use widely.

{\textbf{Inverse tone-mapping}} refers to reconstructing HDR from a single LDR image. The earlier approaches~\cite{banterle2006inverse,masia2009evaluation,akyuz2007hdr,masia2009evaluation} tried to achieve this goal by applying an expand function to the LDR image. Filter based methods were also proposed to deal with both images and videos~\cite{rempel2007ldr2hdr,kovaleski2014high}.
However, these approaches are not user-friendly since they requires to adjust various parameters, which is troublesome for non-expert users to achieve desired results.

{\textbf {Deep learning-based approaches}} have made remarkable achievements in many visual applications like visual recognition~\cite{Long:ICCV2015, Long:CVPR2017, Hua:TPAMI2018, Hu:TIP2021, Hu:arXiv2021}, object detection~\cite{Long:ACCV2014, Islam:CVPR2020}, super resolution~\cite{Zhang:TCSVT2021, Zhang:ICME2020}, image captioning~\cite{Dong:MM2021}, shadow detection and removal~\cite{Ding:ICCV2019, Wei:CGF2019, Zhang:AAAI2020, Zhang:CGF2020, chen:iccv2021}, shadow generation~\cite{ZhangLW19, LiuLZYDX20}, action localization~\cite{Islam:AAAI2021}, trajectory prediction~\cite{Shi:CVPR2021} and so on. Deep learning also provides another way to generate a high-quality HDR image from a single LDR image. Instead of directly producing HDR images by a DCNN model, Endo {\em et al.}~\cite{endo2017deep} developed a DCNN model to generate a stack of bracketed exposure images from a single LDR image. The same idea was adopted by Lee {\em et al.}~\cite{lee2018chain,lee2018deep}, where they used a chaining structure and GAN network to obtain the required bracketed exposure images. However, the HDR images generated by these methods still have a limited dynamic range. 

Recently, directly generating HDR images from a DCNN model is getting more attention. Eilertsen {\em et al.}~\cite{eilertsen2017hdr} developed a virtual camera to generate LDR from HDR datasets, providing enough LDR-HDR pairs to train a model for directly generating HDR image. However, they only predicted the overexposure areas, and the virtual camera was created based on an out-of-date cameras database. Marnerides {\em et al.}~\cite{marnerides2018expandnet} achieved the same goal with a novel network structure by splitting the LDR encoder into local, dilation, and global branches. Kim {\em et al.}~\cite{kim2019deep} learned super-resolution and inverse tone-mapping for UHD HDR applications, but they focus on the final tone-mapped images for display technology and are not concerned with the images in HDR format.  Zhang and Aydın~\cite{DBLP:journals/cgf/ZhangA21} decomposed an input LDR image into a base and detail layer and Liu {\em et al.}~\cite{liu2020single} achieved single image HDR reconstruction by learning to reverse the camera pipeline.  However, the step-by-step strategy leads to the accumulation of errors among the subnetworks. Zhang {\em et al.}~\cite{zhang2017learning} focused on sunlight outdoor lighting estimation on panoramas, but they need to rotate the panorama in advance so that the sun is in the middle of the image. Unlike the existing deep learning methods, we start from the essential properties of HDR images to pre-process the data by calibrating HDR images into a similar scale and introduce the attention mechanism with scale invariance to build our networks. In this way, we propose corresponding HDR reconstruction approaches for both images and panoramas, which can be applied to both HDR display and IBL applications.

\begin{figure}[t]
  \centering
  \mbox{} \hfill
  \includegraphics[width=0.230\textwidth]{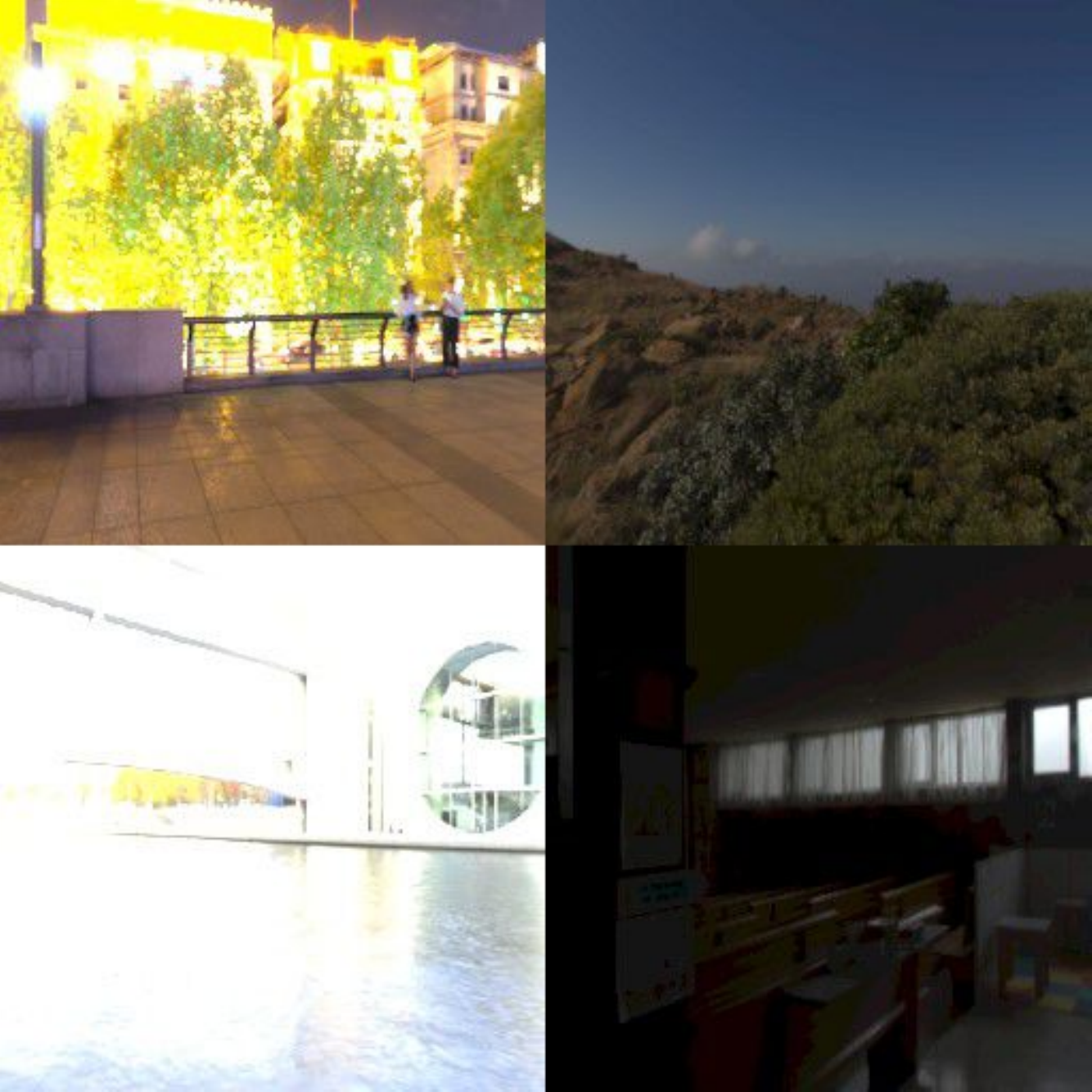}
  \hfill
  \includegraphics[width=0.230\textwidth]{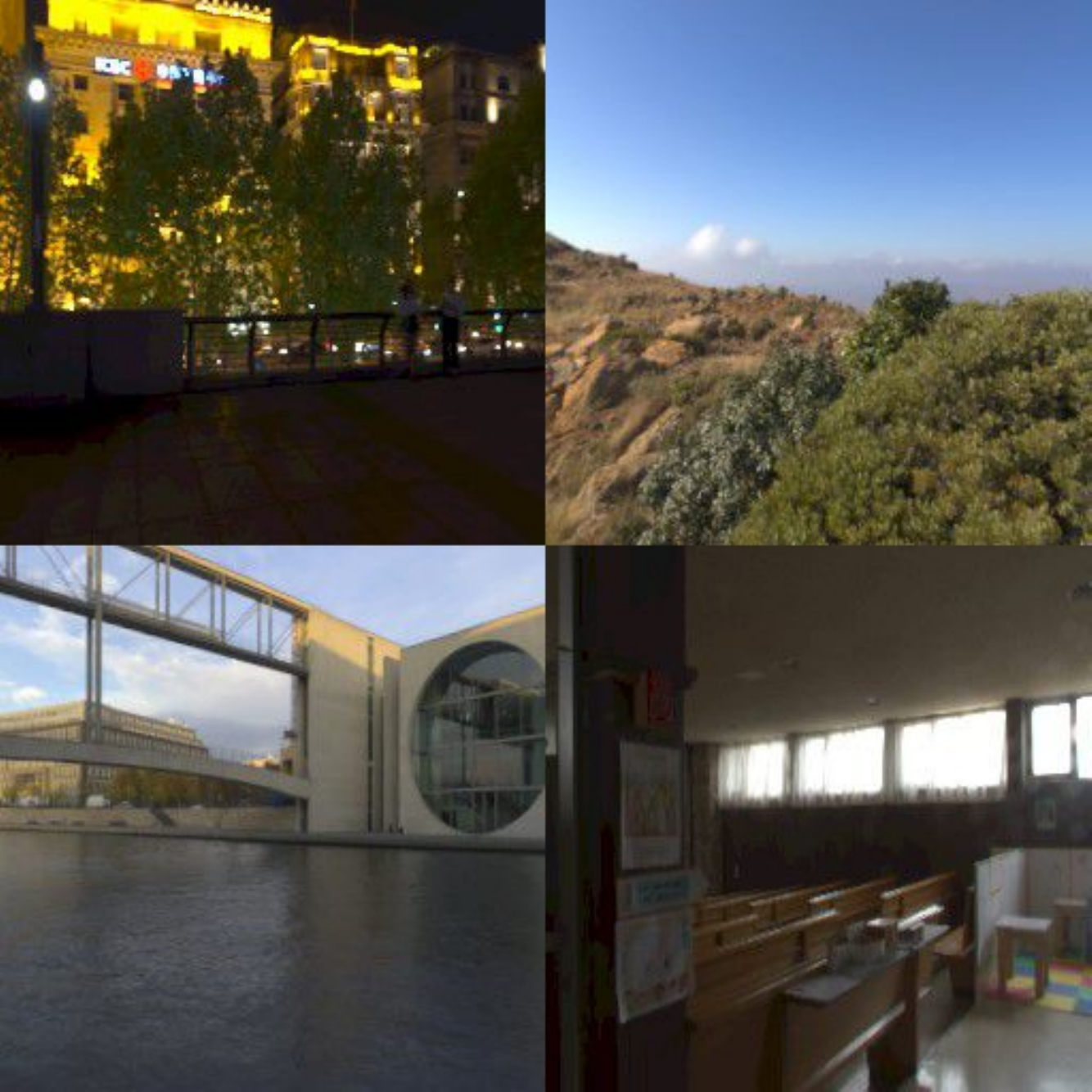}
  \hfill\mbox{}\\
  \hspace{0.03\textwidth}\mbox{(a) Original}\hspace{0.13\textwidth}\mbox{(b) With our method}
  
    \caption{An example of HDR training batch with and without our calibration method. These two batched images are mapped using the LumaninceHDR software in a certain exposure and dynamic range for visualizing. 
    }   
\label{luminance-level}
\vspace{-0.4cm}
\end{figure}

\section{Luminance Attentive Network}
As illustrated in Fig.~\ref{architecture-1}, we propose a novel luminance attentive network named LANet, designed with two streams and fully leverages the auxiliary luminance segmentation for better recovering an HDR image. It has three essential components, {\em i.e.}, HDR calibration with luminance scale invariance, luminance attention module, and luminance scale-invariant loss. We are going to discuss each component as well as the luminance segmentation learning in the following subsections.

\subsection{HDR Calibration with Luminance Scale Invariance}
The main problem of storing HDR images in relative luminance is that the luminance scales may differ. If we use them as ground truth for training a DCNN model, the scale ambiguity will confuse the training process. Therefore, to remove the ambiguity, we propose a novel HDR calibration approach to provide ground truth for training our LANet (see Section~\ref{sec:loss}). The key to our calibration is luminance scale invariance. In the relative luminance domain, we call two HDR images ${{H_1}}$ and ${{H_2}}$ luminance scale-invariant if and only if they are only differed by a positive scale factor $\kappa  \in \mathcal{R}^{+}$ such that ${{H_1}} =  \kappa{{H_2}}$. 
Our goal is to make all HDR images at the same luminance level as much as possible. Since all LDR images are already in the same luminance units and fewer errors in the non-overexposed areas between LDR and HDR images, it is reasonable to choose the LDR version as the standard level to calibrate all of the HDR images. To find the most effective scale factor, for each HDR image ${H}$, we scale it to the same dynamic range with the corresponding normalized LDR image ${I}$ ({\em i.e.}, ${I(x, y)}$ $\in [0, 1]$). The calibrated HDR image ${{\hat{H}}}$ can be obtained as:
\begin{subequations}
\label{eq:Maxwell}
\begin{align}
{{\hat{H}}} &= \frac{S(M \odot {I})}{S(M \odot {H})} \cdot {H} \label{eq-calibrate} \\
S(P) &= \sum\limits_{c, y, x}{P(x, y, c)}, \label{eq-calibrate-sum}\\
M(x, y) &= \begin{cases}
 1,& \text{ if } \frac{1}{3}\sum\limits_{c}{I}(x, y, c) < \tau \\
 0,& \text{ otherwise}
\end{cases} \label{eq-calibrate-m}
\end{align}
\end{subequations}
where $\odot$ denotes element-wise multiplication; $P(x, y, c)$ is a pixel value at position of $(x, y)$ on channel $c$ of an image $P$; $\tau \in [0,1]$ is a threshold used to determine non-overexposed areas, we set $\tau=0.83$ for our experiments which can achieve the best performance in our test. Note that {${I}$} is in the linear RGB space.

As we can see, after applying Eq.~\ref{eq-calibrate}, the HDR image will contain a similar value with the corresponding LDR image in non-overexposed areas, and only the highlight pixels will have a significant difference. As shown in Fig.~\ref{luminance-level}, we can see that images after alignment look more regular at the same specific, while original data have an extensive bias on the luminance scale. What is more, we can apply thresholding segmentation on the calibrated HDR images to obtain pixel-level luminance segmentation labels automatically without requiring any extra manual annotation efforts.

\subsection{Network Architecture and Luminance Attention Module}

The proposed LANet is designed as a multi-task learning framework with two output streams, {\em i.e.}, {\em luminance attention stream} and {\em HDR reconstruction stream}. For each LDR image, the LANet predicts the corresponding HDR image and luminance segmentation mask. We adopt U-Net~\cite{ronneberger2015u} as our backbone structure. Specifically, the encoder follows ResNet50~\cite{he2016deep, he2016identity}, and each of the five residual blocks connects to the decoder by a skip connection. On the decoder side, the following changes have been made.
\begin{itemize}
    \item[$\bullet$] Since a standard deconvolution layer gives checkerboard artifacts~\cite{odena2016deconvolution}, we use a nearest-neighbor upsampling and two convolution layers for each up-sample block.
    \item[$\bullet$] To better keep the distribution of each HDR image, instance normalization~\cite{ulyanov2016instance} is used in our decoder instead of batch normalization~\cite{ioffe2015batch}.
    \item[$\bullet$] luminance attention stream branches out from the middle layer of the decoder and follows by four convolution layers and three upsampling to construct a luminance segmentation mask.
    \item[$\bullet$]  We design a novel luminance attention module (LAM) and incorporate it into the decoder, which brings the luminance attention information from the luminance attention stream to the HDR reconstruction stream.
\end{itemize}

As shown in Fig.~\ref{architecture-1}, LAM is added before the last skip-connection. It takes the last layer of the luminance attention stream and feature maps generated from the second last skip-connection block as input to construct luminance attention feature maps as input to the last skip-connection. With this treatment, the feature maps from the luminance attention stream are transferred into a two-channel attention map through two convolutional and one sigmoid activation layer. The feature maps from the HDR stream also need to go through one convolutional layer. Then, an element-wise multiplication is applied to each channel of the attention map and the generated feature maps from the HDR stream. The outputs are concatenated in channel dimension as an input to another three convolutional layers. The thoughts behind this design are that the attention scheme should give more guidance for both the training and inferring process.

\subsection{Loss Functions\label{sec:loss}}
The overall loss function for LANet is formulated with a luminance scale-invariant loss and a luminance segmentation loss, {\em i.e.},
\begin{equation} \label{eq6}
\mathcal{L} = \mathcal{L}_{SI} + \alpha\mathcal{L}_{SEG},
\end{equation}
where $\alpha$ is a weight hyperparameter to control the trade-off between two losses. Note that $\alpha=0.05$ is used in our experiments.
Inspired by Eilertsen et al.~\cite{eilertsen2017hdr}, the proposed LANet predicts HDR ${{\bar H}}$ in the logarithmic scale, which is better matching how the human visual system reacts to luminance. However, due to ${{\bar H}}$ is still in the relative luminance domain, dealing with the aforementioned scale ambiguity is required. Therefore, we take advantage of luminance scale invariance and define $\mathcal{L}_{SI}$ as a scale-invariant MSE~\cite{grosse2009ground} in logarithmic scale as:
\begin{equation}
\label{siloss-rewrite}
\begin{split}
 \mathcal{L}_{SI}\big({{\bar H}},\ {{\hat{H}}}\big)=&\min_{\kappa \in \mathcal{R}^{+}}\frac{1}{n}\big\|\log(\kappa{{\bar H}}) - \log({{\hat{H}}})\big\|^2 \\
 = &\min_{\kappa \in \mathcal{R}^{+}}\frac{1}{n}\sum_{x, y, c}\Big[\log({\bar H}(x, y,c) + \epsilon) \\
   &- \log({\hat{H}(x, y ,c)} + \epsilon) + \log{\kappa} \Big]^2,
\end{split}
\end{equation}
where $\epsilon$ is a small constant to avoid $\log0$ when calculating logarithms. As we can see, Eq.~\eqref{siloss-rewrite} has a closed-form solution, {\em i.e.},
\begin{equation*}
\log{\kappa} = \frac{1}{n}\sum_{x, y, c}(\log{({\hat{H}(x, y, c)}+\epsilon)} - \log{(\bar H}(x, y, c)+\epsilon)).
\end{equation*}
Therefore, with definition $d(x, y, c) = \log{\bar H(x, y, c)} - \log{\hat H(x, y, c)}$, Eq.~\eqref{siloss-rewrite} can be expressed as:
\begin{equation} \label{eq:L_LI_final}
    \mathcal{L}_{SI}\big({{\bar H}},\ { {\hat{H}}}\big)
   =  \frac{1}{n}\sum_{x, y, c}d(x, y, c)^2 - \frac{1}{n^2}\big(\sum_{x, y, c}d(x, y, c)\big)^2.
\end{equation}
The first component in Eq.~\eqref{eq:L_LI_final} indicates the average of square distances, and the second component refers to the square of the mean distance, ensuring the scale-invariance of the loss function. As we can see, the loss value will not change when the outputs plus a single constant in the logarithmic domain. This means that the network treats HDR images in relative luminance.

The segmentation loss $\mathcal{L}_{SEG}$ is a cross entropy loss, {\em i.e.},
\begin{equation} \label{eq9}
\begin{split}
\mathcal{L}_{SEG}\big(\boldsymbol{m},\ \boldsymbol{\hat{m}}\big) = &-\sum_{x, y, c}\Big[ \hat{m}(x, y, c)\log{m(x, y, c)} \\
&+(1-\hat{m}(x, y, c))\log{(1-m(x, y,c))}\Big],
\end{split}
\end{equation}
where $\boldsymbol{m}$ is a predicted luminance mask; $\boldsymbol{\hat{m}}$ is the ground truth luminance mask, which has the same shape as a reconstructed $\hat{H}$ and initialized with $0$ for all pixel. Then, we set value as follows:
\begin{equation} \label{eq:mask_hat}
\begin{split}
\hat{m}(x, y, c_{idx}) = \begin{cases}
 1 \text{ and } c_{idx}=0, & \text{ if } \hat{H}'(x, y) \leq t_l \\
 1 \text{ and } c_{idx}=1, & \text{ if } t_l < \hat{H}'(x, y) < t_h\\
 1 \text{ and } c_{idx}=2, & \text{ if } \hat{H}'(x, y) \geq t_h,
\end{cases}
\end{split}
\end{equation}
$\hat{H}'(x, y)$ is the channel-wise pixel average value at location $(x, y)$ of ${\hat{H}}$; $t_l$ and $t_h$ are the threshold values to define dimmer and brigher area respectively, which are set to $e^{-5.5}$ and $ e^{0.1}$ in our experiments.


\section{Extension of Luminance Attentive Network for HDR Reconstruction on Panorama}

Regarding HDR images for display and rendering, although both require images with real luminance information, there are two significant differences between them. Firstly, the ways to measure the quality of their application are different. HDR display technology displays the HDR image directly to the user for observation, and the effect depends on the perception of the image itself to the human eye. In contrast, the image-based lighting (IBL) technology uses the HDR image as the scene's light source, and the effect depends on the quality of the rendering results. Secondly, the panorama represents a spherical surface in three-dimensional space, which is invariant to spatial rotation. When it is expanded to equirectangular form, a significant distortion will occur in the upper and lower regions of the image. 

\begin{figure}[b]
  \vspace{-0.3cm}
  \centering
  \includegraphics[width=0.37\linewidth]{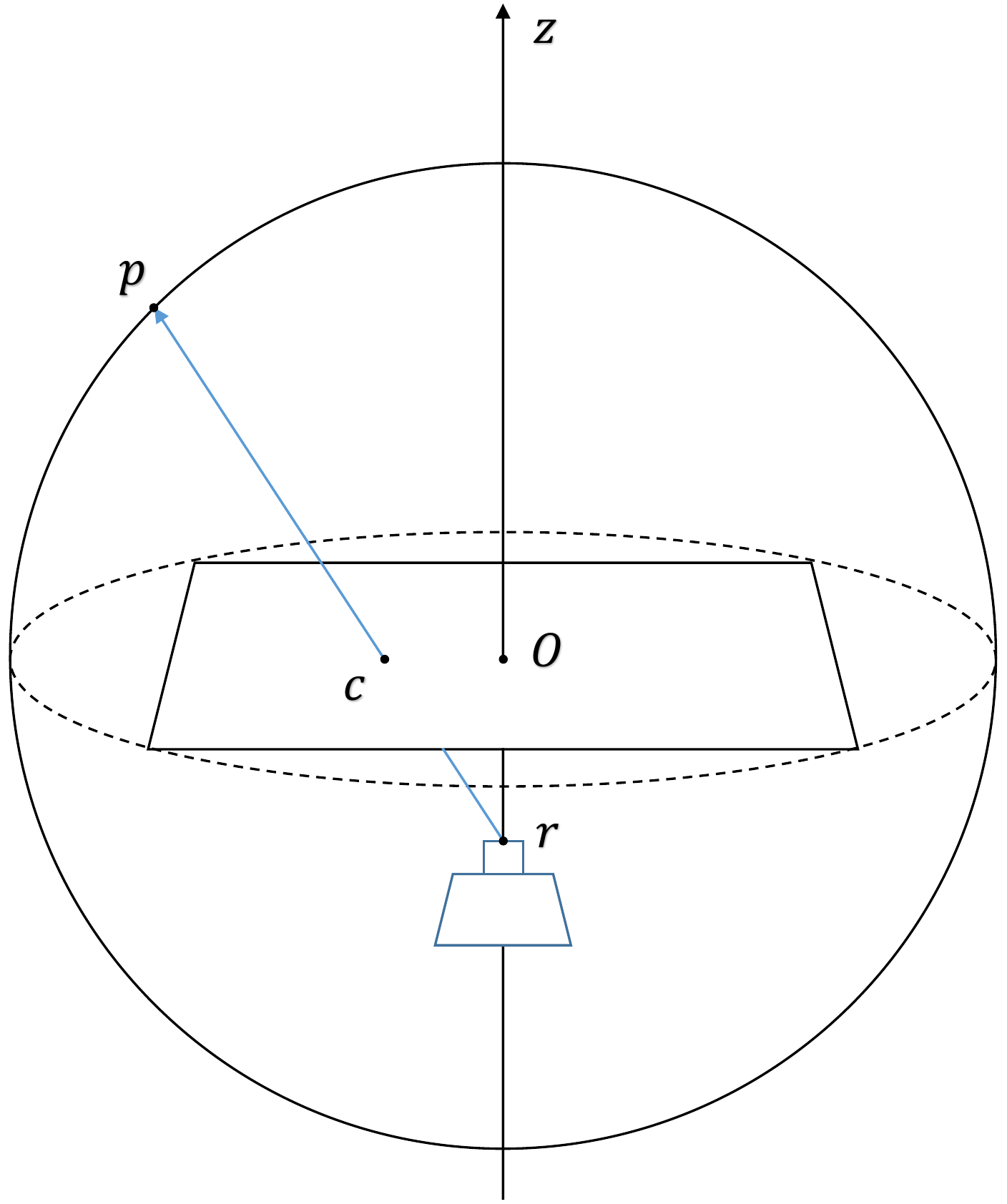}
  \caption{The illustration of the perspective conversion method used in our proposed panoLANet. The entire sphere represents the panorama. The plane at the center of the sphere perpendicular to the $z$-axis is the ceiling-view image. Position $r$ is the camera center of the perspective projection. For any point $p$ on the upper hemispherical surface of the panorama, the intersection point $c$ of $pr$ and the plane is the corresponding projection point.
  }
  \label{P2C-show}
  \vspace{-0.5cm}
\end{figure}

\begin{figure*}[ht!]
  \centering
  \includegraphics[width=0.95\textwidth]{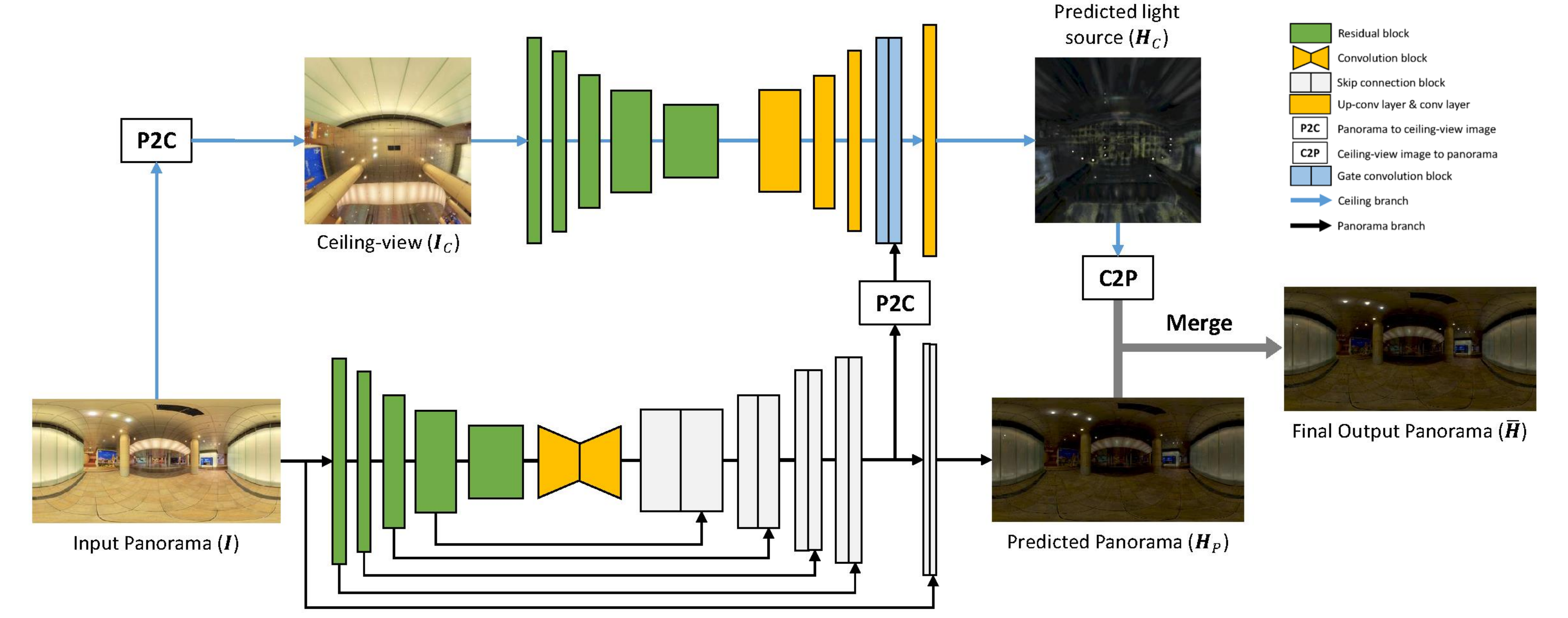}
  \caption{The overview of our panoLANet network, which consists of a ceiling luminance branch and panorama reconstruction branch.
  }
  \label{architecture-2}
  \vspace{-0.4cm}
\end{figure*}

Given these two differences, we extend our proposed LANet and present a novel two-branch network called panoLANet to improve the HDR reconstruction on panoramas for better rendering results. As illustrated in Fig.~\ref{architecture-2}, our panoLANet consists of a ceiling luminance branch and a panorama reconstruction branch. 

Unlike the panorama reconstruction branch, which is used to roughly predict the entire scene's information, the ceiling luminance branch is designed to accurately reconstruct the highlighted area in the upper half of the scene. We use a differentiable perspective transformation method to convert the panoramas and perspective ceiling-view images (P2C and C2P, respectively). Finally, merge them using the mask to obtain the final reconstructed HDR panorama. In the following subsections, we first describe the method of perspective transformation and then explain the detail of the two-branch network and the loss function.

\subsection{P2C and C2P Conversion}
The original panorama represents a two-dimensional spherical surface in three-dimensional space, describing the scene around the shooting center. P2C aims to convert most of the information of the upper hemisphere of this two-dimensional sphere into a plane image through perspective projection. Some previous works directly use the center of the sphere as the center of the projection camera to perform the perspective conversion. However, we want to retain the information in the upper half of the space as much as possible. As shown in Fig.~\ref{P2C-show}, we chose to place the center of the projection camera below the center of the sphere.

\begin{figure}[ht!]
  \centering
  \includegraphics[width=0.98\linewidth]{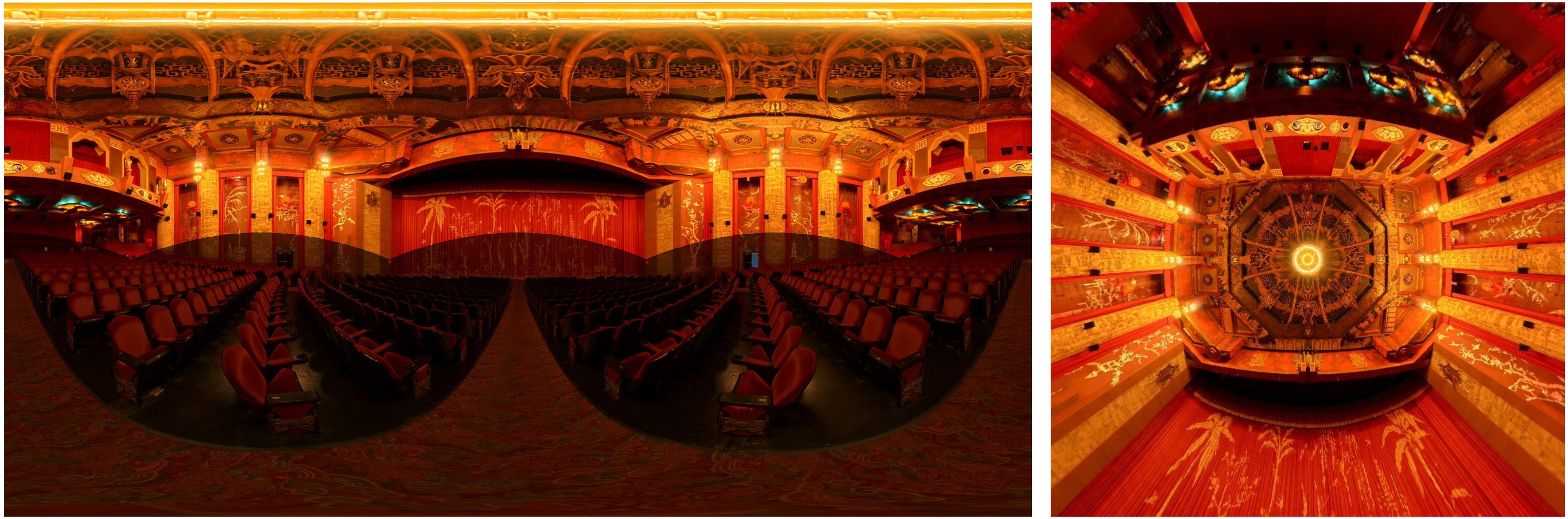}
  \caption{An example result of P2C and C2P conversion: (left) equirectangular panorama, (right) ceiling-view image converted through P2C. Note that when using C2P to convert back to the panorama, the information of the masked area will lose, and therefore we set these uncertain pixels to $0$ in our paper.}
  \label{P2C-example}
  \vspace{-0.5cm}
\end{figure}

Given the field of view of the perspective transformation with the resolution of the panorama and ceiling-view image, we can calculate the position of all pixels of the ceiling-view image on panorama or vice versa, shown as the Fig.~\ref{P2C-show}. Using bilinear interpolation to calculate the final value of each pixel, we get the result of P2C and C2P conversion. Fig.~\ref{P2C-example} shows an example of the conversion of two types of images. It can be seen that the ceiling-view image obtained from P2C can save most of the information of the upper half of the space with less distortion. In addition, by thresholding the ceiling-view image to get the mask of the over-exposed area, we can merge the predicted results of the two branches shown in Fig.~\ref{architecture-2} to get the final HDR panorama.

\subsection{Network and Loss Function}
Looking at Fig.~\ref{architecture-2} with Fig.~\ref{architecture-1}, we can see that panoLANet reuses much of the structure from LANet, which allows us to reduce the training cost through transfer learning. Specifically, the panorama reconstruction branch in panoLANet applies the basic U-Net structure in LANet. The encoder of the ceiling luminance branch also uses the same ResNet50 as the panorama reconstruction branch. We first use the single image data set with an enormous amount of data to train the panorama branch and then use the pretrained model to initialize the panorama reconstruction branch and the encoder of the ceiling luminance branch. Finally, we use the panorama data set to fine-tune the entire panoLANet to get the final training model.

The ceiling luminance branch aims to reconstruct better the HDR of the light source above the space. When we use IBL technology to render, only the upper half of the light affects the rendered object, and the light source is usually above the space rather than below. On the other hand, the light source is the most critical area of the image-based lighting, so the results rendered from LDR panoramas often look unrealistic. We convert the panorama to a ceiling-view image and then build an encoder-decoder network to focus on reconstructing the light source. Specifically, to avoid the loss of information caused by the downsampling operation, we use a skip connection to share the features from the panorama reconstruction branch to the ceiling luminance branch. Based on the feature from skip connection, we use the gated convolution~\cite{yu2019free} to achieve local attention, which is called gate convolution block on Fig.~\ref{architecture-2}, so that the network can focus more on the over-exposed areas.

Our goal is to recover HDR information from the LDR panorama for image-based lighting. In order to get the final HDR reconstruction result, we need to merge the prediction results of the two branches. Given the prediction of the ceiling luminance branch ${H}_C$ and prediction of the panorama branch ${H}_P$, the merged final output HDR panorama ${{\bar H}}$ could be written as
\begin{equation} \label{eq:merge}
    {{\bar H}} = {m}_p\odot\mathcal{F}_{c2p}\left({H}_C\right) + (1-{m}_p)\odot{H}_P,
\end{equation}
where $\mathcal{F}_{c2p}$ is the C2P conversion and $\odot$ denotes element-wise multiplication. The ${m}_p$ is the mask generated by ceiling-view input ${I}_C$ as
\begin{equation} \label{eq:merge-mask}
    {m}_p = \frac{\max(0, \text{avg}_c(\mathcal{F}_{c2p}({I}_C)) - \tau)}{1 - \tau},
\end{equation}
where $\text{avg}_c$ means average on the image channels and $\tau=0.13$ is a threshold to generate the mask. Note that ${I}_C$ is in linear RGB space.

The loss function in panoLANet is no longer a scale-invariant loss, because the scale-invariant loss described in Eq.~\ref{eq:L_LI_final} requires that each pixel of the image contributes equally to the loss. Here we use two loss functions $\mathcal{L}_{high}$ and $\mathcal{L}_{low}$ to describe the loss of over-exposed areas and non-overexposed areas, respectively. We define ${d} = \log{{\bar H}} - \log{{\hat H}}$, then the final loss function for panoLANet is
\begin{equation} \label{eq:loss-pano}
\begin{split}
    \mathcal{L}\big({{\bar H}},\ {{\hat{H}}}\big) &= \beta_{1}\mathcal{L}_{high}\big({{\bar H}},\ {{\hat{H}}}\big) + \beta_{2}\mathcal{L}_{low}\big({{\bar H}},\ {{\hat{H}}}\big) \\
    &= \beta_{1}\big|{m}_{p} \odot {d}\big|^2 + \beta_{2}\big|(1 - {m}_{p}) \odot {d}\big|^2,
\end{split}
\end{equation}
where $\beta_1, \beta_2$ is weight hyperparameters to control the trade-off between two losses. We set $\beta_1=0.2, \beta_2=0.01$ in our experiments.

\section{Experiments}\label{experiment}
To validate our proposed method, we collect several public HDR datasets shown in Table~\ref{datalist} and conduct various experiments on these datasets. For single image HDR reconstruction of our LANet, we pick out the HDR-Eye dataset~\cite{nemoto2015visual} from the collected dataset and use the remaining HDR data to generate training and test data. Since the HDR-Eye dataset has both real LDR and HDR data, we use this dataset as an additional comparison dataset with the previous work to show the applicability of our method. For panorama HDR reconstruction, we use the sIBL~\cite{hdrlabs} dataset with both LDR and HDR panorama data as our test data.

\begin{table}[ht!]
\centering
\caption{The list of HDR datasets}
\label{datalist}
\begin{tabular}{|c|l|c|}
    \hline
    Type & Dataset Name & Number \\
    \hline
    \multirow{4}{*}{Pano} & Laval Indoor HDR Dataset~\cite{gardner2017learning} & 2233 \\ 
    ~ & Laval Outdoor HDR Dataset~\cite{Hold-Geoffroy_2019_CVPR} & 205 \\ 
    ~ & HDRI Haven~\cite{HDRIhaven} & 322 \\ 
    ~ & sIBL~\cite{hdrlabs} & 79 \\ 
    \hline
    \multirow{4}{*}{Img} & HDR Photographic Survey~\cite{hdrsurvey} & 105 \\ 
    ~ & Funt et al. HDR Dataset~\cite{FuntHDR} & 105 \\ 
    ~ & Stanford HDR Data~\cite{xiao2002high} & 88 \\ 
    ~ & Ward~\cite{wardHDR} & 33 \\ 
    ~ & HDR-Eye~\cite{nemoto2015visual} & 42 \\ 
    \hline
    Video & LiU HDRv~\cite{KGBU13,kronander2014unified,eilertsen2016evaluation} & 10 \\ 
    \hline
\end{tabular}
\vspace{-0.5cm}
\end{table}

\subsection{Implementation Detail}
We calibrate all HDR images using Eq.~\ref{eq-calibrate} and resize the training pairs to the size of $256\times256$ for general images data and $512\times256$ for panoramas data. We implement our proposed LANet and panoLANet in TensorFlow and adopt ADAM optimizer~\cite{kingma2014adam} to train the model with one NVIDIA RTX 2080Ti GPU. The training takes 100K iterations in total with a batch size of 16 for LANet and takes 100K iterations for pretraining with a batch size of 16 and 35k iterations for fine-tuning with a batch size of 4 for panoLANet. Regarding the learning rate, we set the initial learning rate as $4e{-5}$ and update it with a step decay schedule. Specifically, the learning rate is dropped $0.8$ times for every 5000 iterations.

\subsection{Datasets and Metrics}
\begin{figure}[t]
  \centering
  \mbox{} \hfill
  \includegraphics[width=0.225\textwidth]{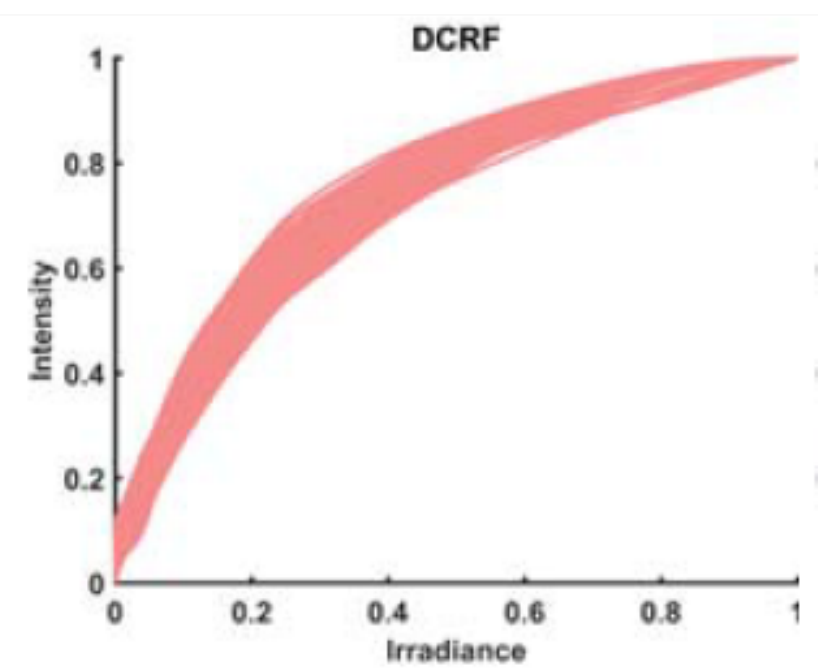}
  \hfill
  \includegraphics[width=0.225\textwidth]{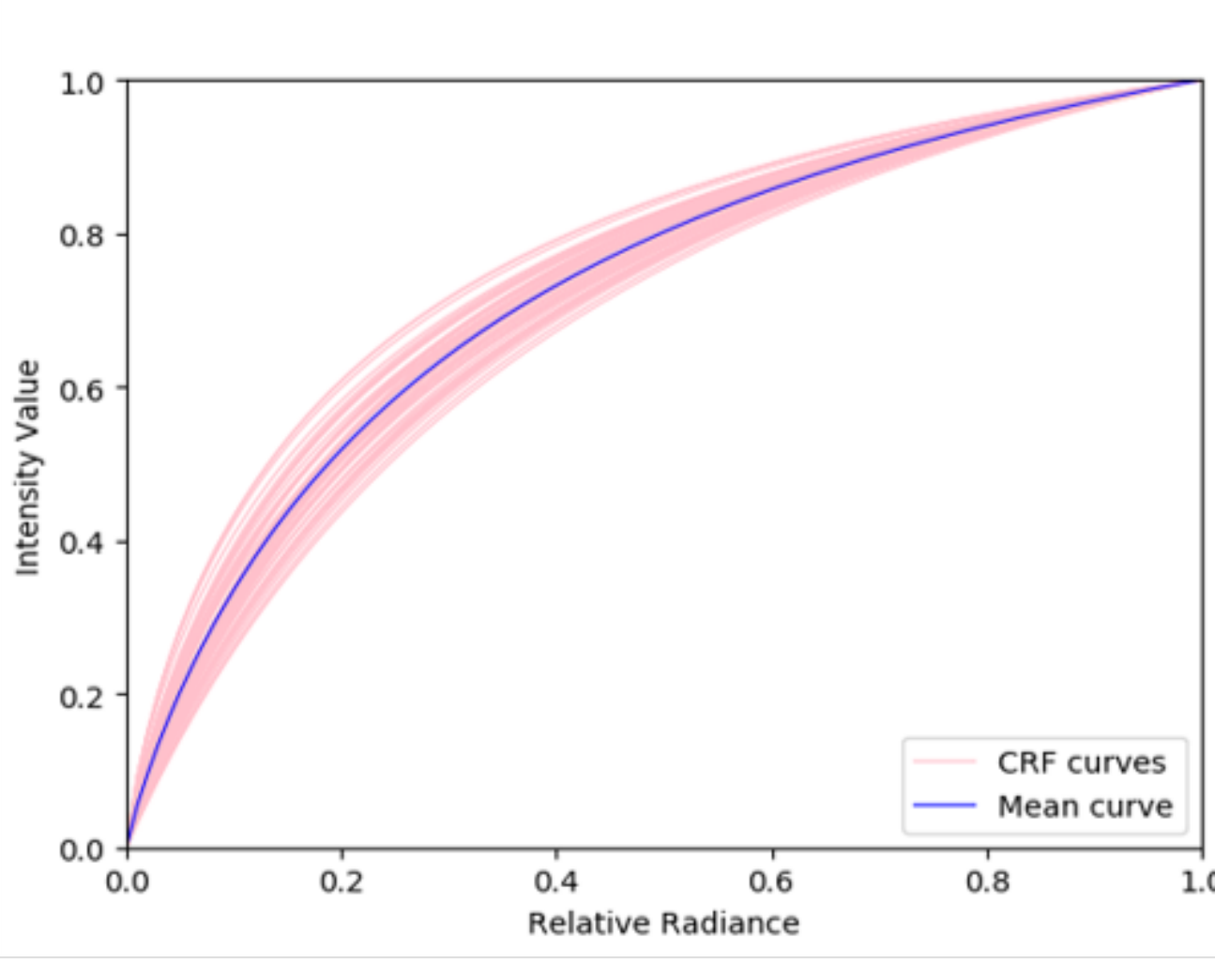}
  \hfill\mbox{}\\
  \mbox{(a) Modern CRF curves~\cite{chen2019analyzing}}\hspace{0.01\textwidth}\mbox{(b) With our method}
\caption{The modern CRF curves collected by Chen et al. and our random curves. We simulate the CRF curves through a sigmoid function that parameters are set from a certain range.}
\label{CRF-curve}
\end{figure}

The entire HDR raw data we collect contains 2,839 HDR panoramas, 373 single HDR images and 10 HDR videos. For each panorama, we crop images to cover six equidistant azimuths in the horizontal viewing angle and three equidistant azimuths in the 45-degree elevation angle (except outdoor scenes). Each cropped image has an aspect ratio of $4:3$. We crop images in the center with two aspect ratios for every single HDR image, either $4:3$ or $3:4$. For each HDR video, we select the suitable frames to be included in our training data. In this way, we get 25,308 HDR images in total for experiments. Note that most HDR datasets have only HDR images. We need to generate the synthetic LDR images from HDR data. To simulate the modern cameras, we complete our virtual camera by following steps:
\begin{enumerate}
\item We randomly set a dynamic range for the virtual camera from $9.6 EV$ to $14.8 EV$ based on the popular Digital Single Lens Reflex (DSLR) cameras~\cite{dxomark}.
\item According to the dynamic range, we use a mean-value auto-exposure algorithm to find an exposure that makes the mean value of the mapped LDR image approximately to middle gray. After that, we obtain the mapped linear LDR image in which all pixels are in the range $[0,1]$.
\item We apply approximate Camera Response Function (CRF) curves to the linear images, which are in the same form as Eilertsen {\em et al.}~\cite{eilertsen2017hdr},
\begin{equation} \label{eq4}
f\big(H_{l,c}\big) = (1+\sigma)\frac{H_{l,c}^n}{H_{l,c}^n + \sigma},
\end{equation}
while we here set the $\sigma$ in range $[0.3,0.5]$ and $n$ in range$[0.8,1.0]$ to fit a modern CRF database.~\cite{chen2019analyzing}). The modern CRF and our approximate curves are shown in Fig.~\ref{CRF-curve}.
\end{enumerate}

\begin{table}[ht!]
  \caption{Quantitative Results of Ablation Study on LANet. $\mathbf{N}$ indicates that the data has been normalized while $\mathbf{C}$ indicates calibrated. }
  \label{ablation-table}
  \centering
  \begin{tabular}{l|cccc}
    \hline
    Method & PU-PSNR$\uparrow$ & PU-SSIM$\uparrow$ & $Q_H\uparrow$ & siMSE$\downarrow$ \\
    \hline
    U-Net+$L_2(\mathbf{N})$	 & $33.09$ & $0.922$ & $38.31$ & $7.93$ \\
    U-Net+$L_2(\mathbf{C})$	   & $35.04$ & $0.963$ & $41.63$ & $6.70$ \\
    U-Net+ $L_{SI}$      & $35.37$ & $0.966$ & $42.58$ & $6.95$ \\
    \hline
    w/o $L_{SI}(\mathbf{N})$   & $34.31$ & $0.952$ & $37.33$ & $6.34$ \\
    w/o $L_{SI}(\mathbf{C})$      & $35.24$ & $0.969$ & $40.31$ & $6.30$ \\
    \hline
    w/o LAM           & $35.27$ & $0.965$ & $42.38$ & $7.12$ \\
    \hline
    w/o Seg   & $35.65$ & $0.972$ & $\mathbf{42.59}$ & $5.99$ \\
    w/ LDR-Seg    & $35.70$ & $0.971$ & $42.33$ & $6.22$ \\
    w/ HDR-Seg    & $\mathbf{35.79}$ & $\mathbf{0.973}$ & $42.53$ & $\mathbf{5.62}$ \\
    \hline
  \end{tabular}
  \vspace{-0.5cm}
\end{table}

\begin{figure*}[ht!]
  \centering
  \includegraphics[width=0.95\textwidth]{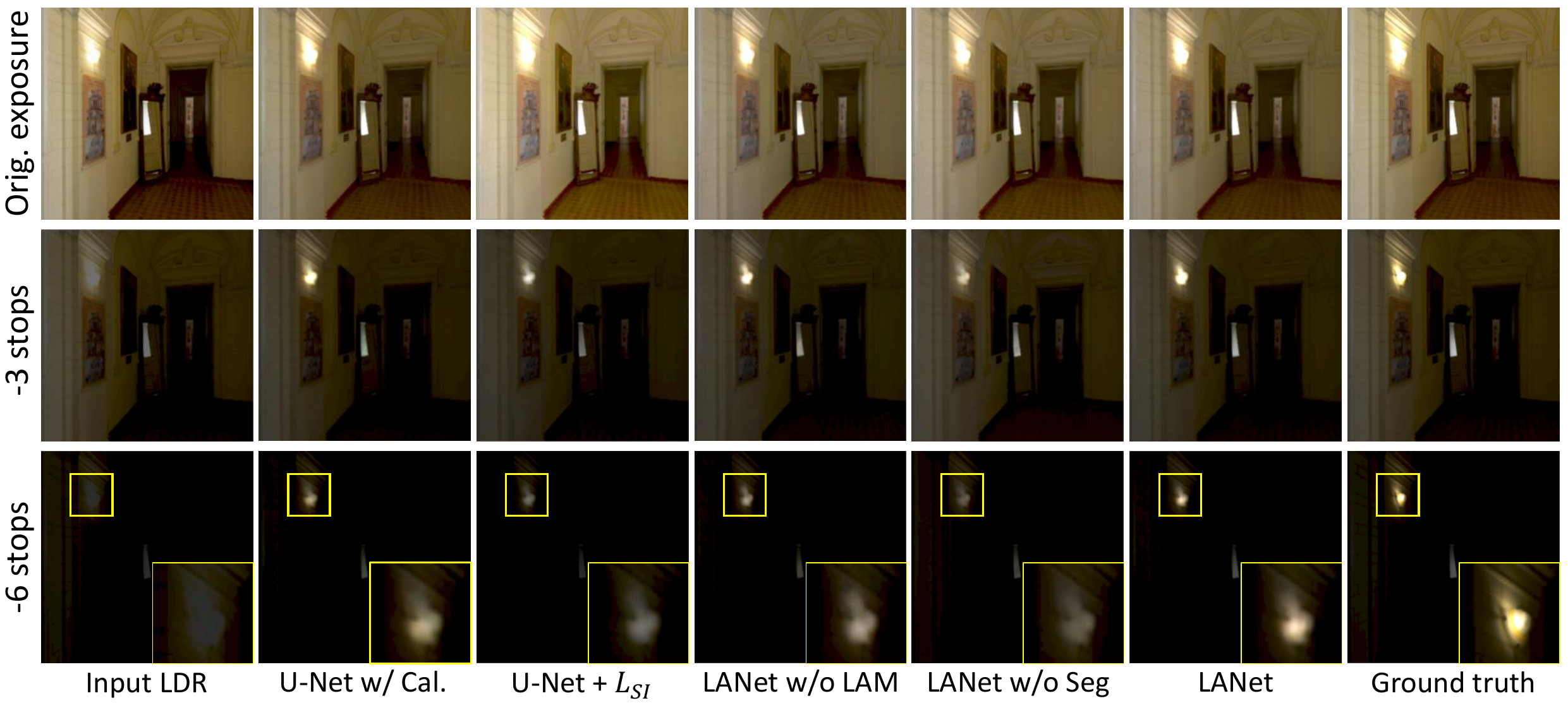}
  \caption{An example of the ablation models. We show the same images in different reduced exposure. For each image, we use the same dynamic range and a manually adjusted exposure with LuminanceHDR software on the original image for the best visual comparison.}
  \label{ablation-figure1}
  \vspace{-0.4cm}
\end{figure*}

The final mapped LDR images are scaled to $0-255$ in integer to fit the available 8-bits LDR images. For each HDR image, we generate one LDR image using the above virtual camera and the other one with Display Adaptive TMO~\cite{mantiuk2008display} to fit the images that may be post-processed. We finally got a total of 50616 image pairs in our experiment.

For experiments on general images, we randomly select $90\%$ image pairs for training. For the remaining 10\% image pairs, we discard all pairs obtained from Laval HDR Dataset~\cite{gardner2017learning,Hold-Geoffroy_2019_CVPR} to avoid testing cases over-concentrated in a majority dataset and use the rest 802 pairs for evaluation. We also evaluate comparison performance on the HDR-Eye dataset~\cite{nemoto2015visual} which contains 46 pairs of ground truth HDR images and LDR images captured from several different cameras. Note that we only have 42 pairs for evaluation because the LDR images have large black regions in the first four pairs.

We remove the data generated from the sIBL dataset for experiments on panoramas and randomly select $90\%$ image pairs for pretraining. Then we use the corresponding panoramas of the pretrained data to fine-tune the panoLANet model. We evaluate our panoLANet on the sIBL dataset~\cite{hdrlabs} which contains 79 pairs of ground truth HDR panoramas and corresponding LDR panoramas. Note that we conduct the evaluation on rendering results rather than panorama itself, which will be described in detail in Sec.~\ref{evaluate-pano}.

Regarding the metrics, we use the Q score of HDR-VDP-2~\cite{narwaria2015vdp} as $Q_H$ score, scale-invariant MSE (siMSE)~\cite{eigen2014depth}, perceptual uniformity encoded PSNR (PU-PSNR) and SSIM (PU-SSIM)~\cite{aydin2008extending}, to evaluate the quality of output HDR images in HDR domain. 
Since the HDR-VDP-2, PU-PSNR and PU-SSIM metrics are display-referred metrics applicable for absolute luminance in $cd/m^2$, we need to adjust the luminance level to a certain absolute luminance. We first scale the predicted HDR images to the ground truth HDR images, and then we adjust their luminance level approximately to where the maximum value of the corresponding LDR image is $255cd/m^2$. For siMSE metric, we need not change results when comparing this metric because it is completely scale-invariant. As the default configuration for HDR-VDP-2 metric, color encoding is set as ``rgb-bt.709" for HDR evaluation, 24-inch display, $1920\times1080$ resolution and viewing distance of 1 meter.

\subsection{Ablation Study on LANet}
To evaluate some components of our proposed LANet, we design a series of variants as follows:
\begin{itemize}
    \item[$\bullet$] U-Net + $L_2$: remove attention stream and LAM and use $L_2$ loss for training.
    \item[$\bullet$] U-Net + $L_{SI}$: remove attention stream and LAM and use $L_{SI}$ loss for training.
    \item[$\bullet$] w/o $L_{SI}$: use $L_2$ loss rather than $L_{SI}$, and without using supervised information for luminance segmentation.
    \item[$\bullet$] w/o LAM: remove LAM to make luminance segmentation and HDR reconstruction separately.
    \item[$\bullet$] w/o Seg: Full LANet model, but without using supervised information for luminance segmentation.
    \item[$\bullet$] w/ LDR-Seg: use the lumninance segmentation supervised information calculated based on the input LDR images rather than calibrated HDR images.
    \item[$\bullet$] w/ HDR-Seg: use the lumninance segmentation supervised information calculated based on the calibrated HDR images.
\end{itemize}

Meanwhile, we evaluate the general maximum normalization method and our calibration method on ``U-Net + $L_2$" and ``LANet + $L_2$" which can take HDR data with different luminance scales as ground-truth for training. Note that there is no difference for these two processing when using $L_{SI}$. 

We train the above mentioned variants with our proposed LANet on the same training data and evaluate their results on our self-collected 802 pairs of LDR-HDR images. The results are summarized in Table~\ref{ablation-table}, we use the $Q_{H}$ score as the value of HDR-VDP-2 metric and the results of siMSE metric are in $10^{-2}$ units. From the results, we can observe: (1) for both U-Net+$L_2$ and LANet+$L_2$, the performance with calibrated HDR as ground-truth for training is better than taking general normalized HDR as ground-truth; 
(2) with $L_{SI}$ loss, the performances of U-Net and LANet are better than with $L_2$ loss, which verifies the effectiveness of scale invariance in HDR; (3) with $L_{SI}$ loss, all of the LANet model performs better than U-Net, which suggests that LAM works with or without using luminance segmentation labels for supervision; (4) LANet shows superiority when compared with LANet w/o LAM, which demonstrates the efficacy of LAM with luminance attention stream; (5) with LDR-Seg to replace HDR-Seg, LANet w/ LDR-Seg achieves lower performance than LANet, which indirectly demonstrate the effectiveness of luminance segmentation labels obtained from calibrated HDR rather than input LDR images; and (6) LANet w/ HDR-Seg achieves the best performance in all the metrics except $Q_H$ score, which shows that using luminance segmentation supervision can improve the performance. Because adding the segmentation supervision does not require additional calculations, we use the LANet w/ HDR-Seg as the final model of LANet.

\begin{figure}[ht!]
  \centering
  \includegraphics[width=0.98\linewidth]{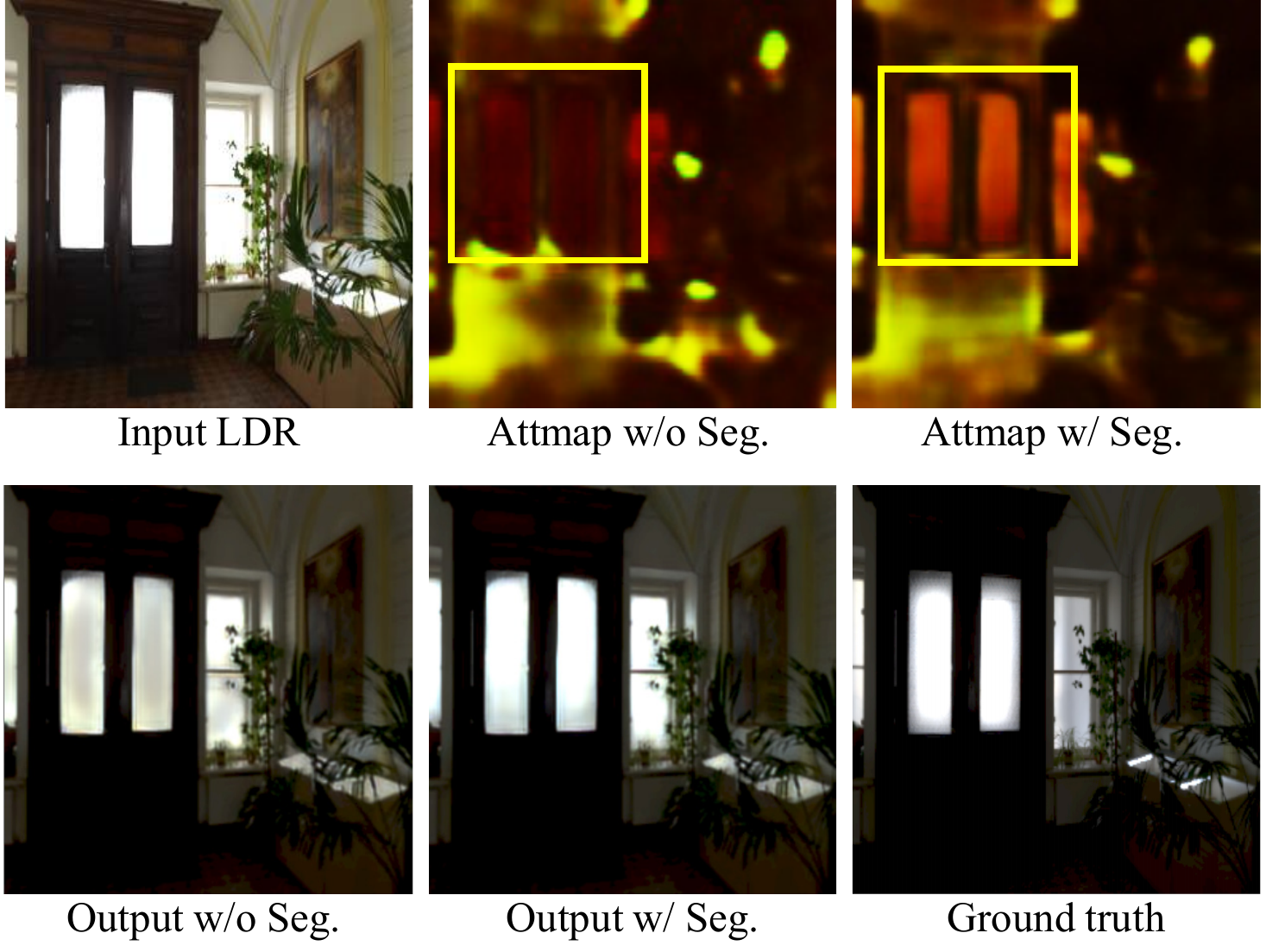}
  \caption{An example of attention map generated with and without segmentation learning. Two attention maps are applied a same color boost for visualizing.}
  \label{ablation-ofigure-att}
   \vspace{-0.4cm}
\end{figure}

We also provide the visual comparison results in Fig.~\ref{ablation-figure1}, from which we can see that our proposed LANet recovers the best details and the generated HDR image looks more realistic and more convincing especially viewing under low exposure. To better verify our luminance attention stream and the well-designed LAM, we visualize the attention maps with and without the luminance attention stream as well as the corresponding reconstructed HDR images in Fig.~\ref{ablation-ofigure-att}. Obviously, with the auxiliary luminance attention stream and LAM, our proposed LANet can recovery better HDR images.

\begin{table*}[ht!]
  \caption{Quantitative Results on HDR Domain}
  \label{compare-table}
  \centering
  \begin{tabular}{l|cccc|cccc}
    \hline
    \multirow{2}{*}{Method} & \multicolumn{4}{c|}{802 testing pairs} & \multicolumn{4}{c}{HDR-Eye dataset} \\
      	& \ PU-PSNR$\uparrow$ \ & \ PU-SSIM$\uparrow$ \ & \ $Q_{H}\uparrow$ \ & \ siMSE$\downarrow$ & \ PU-PSNR$\uparrow$ \ & \ PU-SSIM$\uparrow$ \ & \ $Q_{H}\uparrow$ \ & \ siMSE$\downarrow$ \ \\
    \hline
    KOEO~\cite{kovaleski2014high} 			   & $26.54$ & $0.870$ & $36.34$ & $2.76$  & $18.91$ & $0.502$ & $29.26$ & $8.95$\\
    HDR-CNN~\cite{eilertsen2017hdr}
                                                & $28.89$ & $0.906$ & $39.97$ & $1.97$ & $20.39$ & $0.572$ & $32.29$ & $6.74$\\
    DrTMO~\cite{endo2017deep} 	
                                                & $32.12$ & $0.935$ & $39.39$ & $1.55$ & $24.90$ & $0.781$ & $32.34$ & $2.94$\\
    Expand-Net~\cite{marnerides2018expandnet}  & $27.30$ & $0.862$ & $ 37.38$ & $2.63$ & $22.29$ & $0.675$ & $30.81$ & $5.02$\\
    
    Santos {\em et al.}~\cite{DBLP:journals/tog/SantosRK20}  & $27.88$ & $0.886$ & $ 39.99$ & $2.22$ & $20.35$ & $0.568$ & $31.70$ & $0.652$\\
    
    Liu {\em et al.}~\cite{liu2020single}  & $31.21$ & $0.928$ & $ 39.97$ & $1.43$ & $25.57$ & $0.812$ & $34.62$ & $3.01$\\
    
    \hline
    LANet   & $ \mathbf{35.79} $ & $\mathbf{0.973}$ & $\mathbf{42.53}$ & $\mathbf{0.56}$ & $ \mathbf{26.08} $ & $\mathbf{0.816}$ & $\mathbf{35.63}$ & $\mathbf{2.46}$\\
    \hline
  \end{tabular}
  \vspace{-0.25cm}
\end{table*}

\begin{figure*}[ht!]
  \centering
  \includegraphics[width=0.95\textwidth]{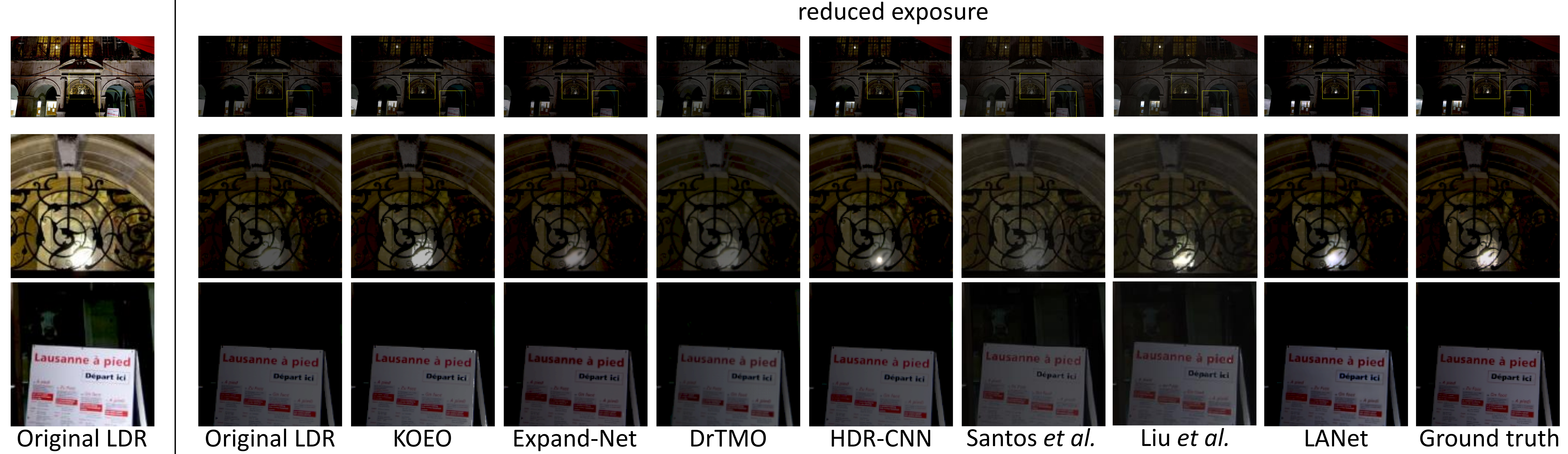} \label{compare_a} \\
  \includegraphics[width=0.95\textwidth]{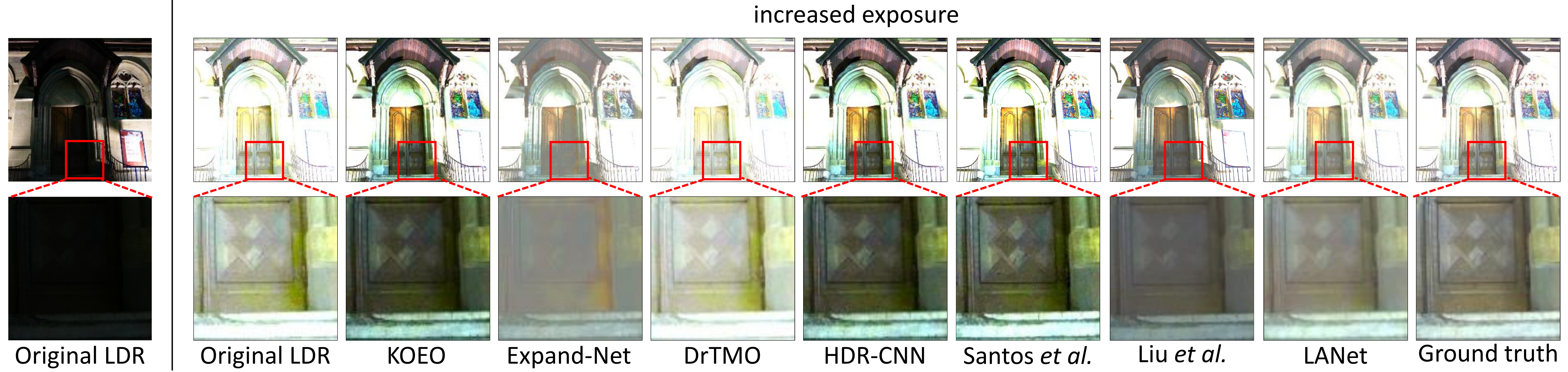}\\ \label{compare_b}
  \caption{Qualitative results and comparison under different mapping approaches. The comparison on predicted HDR is under the same dynamic range and an approximate exposure for best visual comparison.}
  \label{compare-figure-2}
   \vspace{-0.3cm}
\end{figure*}

\subsection{Comparing LANet with State-of-the-Art}

We compare our LANet with one traditional method KOEO~\cite{kovaleski2014high}, and five deep learning-based methods, {\em i.e.},  HDR-CNN~\cite{eilertsen2017hdr}, DrTMO~\cite{endo2017deep}, Expand-Net~\cite{marnerides2018expandnet}, Santos {\em et al.}~\cite{DBLP:journals/tog/SantosRK20} and Liu {\em et al.}~\cite{liu2020single} on both our self-collected 802 pairs of testing data and the HDR-Eye HDR dataset~\cite{nemoto2015visual} for quantitative evaluation. The results are summarized in Table~\ref{compare-table}, where $Q_H$ is the $Q$ score of HDR-VDP-2 metric, and the results of siMSE metric are in $10^{-1}$ units. From the table, we can observe that our proposed LANet outperforms the competing methods in terms of $Q_{H}$, PU-PSNR, PU-SSIM, and siMSE. To further explain the outperformance of our proposed LANet, we provide some visualization results in Fig.~\ref{compare-figure-2}. The figure shows that our method can reconstruct a more realistic dynamic range in over-exposed areas and recover the detail of luminance change at non-overexposed areas. 


\begin{figure}[ht!]
  \centering
  \includegraphics[width=0.47\textwidth]{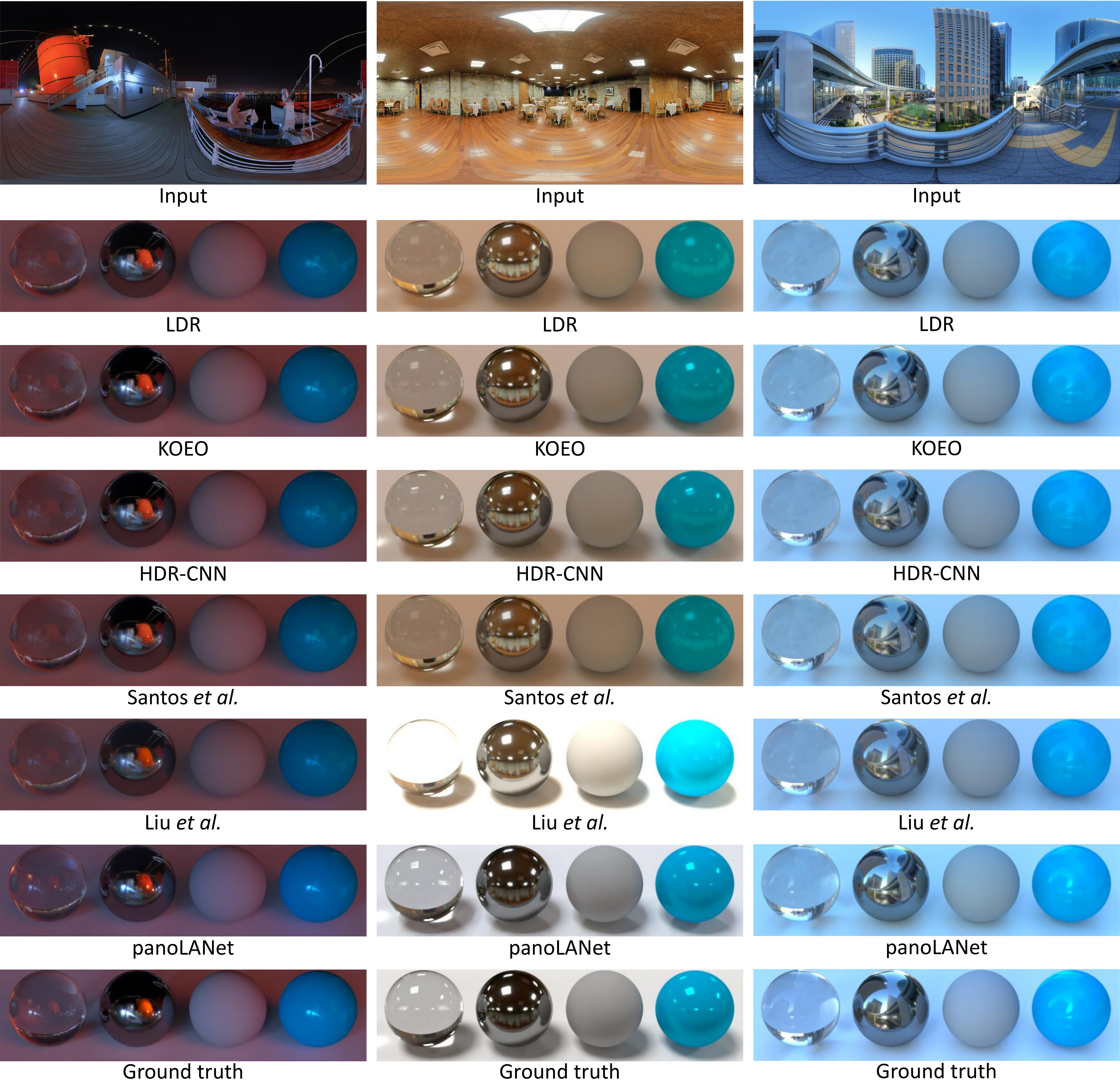}
  \caption{Comparisons on rendering results of predicted panoramas in different scenes. The materials of the four balls from left to right are: glass, glossy, diffuse and mixed diffuse with glossy. 
  }
  \label{compare-pano-2}
  \vspace{-0.5cm}
\end{figure}

\subsection{Application on Image-based Lighting \label{evaluate-pano}} 
This section evaluates our panoLANet by applying the reconstructed HDR panoramas to IBL with physically-based rendering. Unlike general images used for display, HDR panoramas mainly provide real environment lighting for physically-based rendering (PBR). Therefore, it is inaccurate to directly evaluate the quality of the panorama by calculating the metrics on images, while we need to evaluate the rendering results rendered by panoramas.

However, for common HDR panoramas, an important issue is that they are not on the same luminance unit. This makes us render the virtual objects by manually adjusting the light intensity of the panorama. When we perform a quantitative evaluation of a large amount of data, such a manual adjustment will take a long time. What is more, manually adjusting the light intensity is subjective, which will cause the final evaluation results can not be trusted.

Here we can directly use the proposed HDR calibration method to avoid this problem. We calibrate them for each HDR panorama (predicted and ground truth) based on the corresponding LDR panorama in the sIBL dataset. Then we use the calibrated HDR panorama to render a virtual scene under the same rendering configuration. Finally, we make a quantitative comparison of the rendered images to complete the evaluation of our proposed method. Specifically, we use the Blender software with the Cycle rendering engine as the rendering tool for our experiments. The virtual scene we rendered for evaluation consists of four balls with different materials above a diffuse plane, which can be seen in Fig.~\ref{compare-pano-2}. 

\begin{table}[ht!]
  \caption{Comparison of Rendering Results}
  \label{ablation-pano}
  \centering
  \begin{tabular}{l|cccc}
    \hline
    Method &  PSNR$\uparrow$  &  SSIM$\uparrow$  &  VDP$\uparrow$  &  MSE$\downarrow$  \\
    \hline
    KOEO~\cite{kovaleski2014high} 	& $24.10$ & $0.847$ & $38.55$ & $3.69$  \\
    HDR-CNN~\cite{eilertsen2017hdr} & $23.00$ & $0.831$ & $36.14$ & $4.22$\\
    Santos {\em et al.}~\cite{DBLP:journals/tog/SantosRK20} & $22.91$ & $0.827$ & $35.26$ & $4.73$\\
    Liu {\em et al.}~\cite{liu2020single} & $23.81$ & $0.861$ & $37.77$ & $3.05$\\
    
    \hline
    U-Net w/ Norm	  &  $25.13$ & $0.847$ & $38.48$ & $1.81$ \\
    U-Net w/ Cal	  &  $25.22$ & $0.892$ & $40.43$ & $2.04$ \\
    U-Net w/ $L_{SI}$ &  $26.07$ & $0.910$ & $40.68$ & $1.81$ \\
    panoLANet w/o Att &  $24.26$ & $0.863$ & $37.84$ & $3.91$ \\
    LANet             &  $26.20$ & $0.901$ & $41.31$ & $1.71$ \\
    \hline
    panoLANet       &  $\mathbf{26.79} $ & $\mathbf{0.926}$ & $\mathbf{41.66}$ & $\mathbf{1.35}$ \\
    \hline
  \end{tabular}
\end{table}

Since DrTMO~\cite{endo2017deep} and Expand-Net~\cite{marnerides2018expandnet} aimed to reconstruct HDR for display, we compare our panoLANet with KOEO~\cite{kovaleski2014high}, HDR-CNN~\cite{eilertsen2017hdr}, Santos {\em et al.}~\cite{DBLP:journals/tog/SantosRK20} and Liu {\em et al.}~\cite{liu2020single} only. In addition, to evaluate the components of our panoLANet, we design a series of variants as follows:
\begin{itemize}
    \item[$\bullet$] U-Net w/ Norm: use basic U-Net with the general maximum normalization method. 
    \item[$\bullet$] U-Net w/ Cal: use basic U-Net with our calibration method.
    \item[$\bullet$] U-Net w/ $L_{SI}$: use basic U-Net with scale-invariant loss.
    \item[$\bullet$] LANet: use the same structure as LANet, but applying fine-tune on the panoramas dataset.
    \item[$\bullet$] panoLANet w/o Att: use the full panoLANet without skip connection and gated convolution layer.
\end{itemize}
We train the above mentioned variants with panoLANet on the same training data and evaluate their results on the sIBL dataset~\cite{hdrlabs}. The results are summarized in Table~\ref{ablation-pano} and the results of siMSE metric are in $10^{-3}$ units. From the results, we can see that our panoLANet gets the best results in all the metrics, especially the result of SSIM is as high as 0.926. In addition, the HDR calibration algorithm is also superior to the general maximum normalization method in the HDR reconstruction of panorama, which again indicates the advantages of our calibration method. 

In order to demonstrate the advantages of panoLANet more clearly, we show the comparison results of our method with the previous methods in three different scenes: night scene, indoor and outdoor scenes. As can be seen from the qualitative comparison results in Fig.~\ref{compare-pano-2}, our method estimates the position and brightness of the light source more accurately, which is reflected in the more realistic highlight position and brightness of the render results. In addition, the $\mathcal{L}_{SI}$ is adopted to promote a more accurate ratio relationship between the intensity of the light source and the background, which can be observed from the results in the second column of Fig.~\ref{compare-pano-2}. 

\subsection{Discussion}
\begin{figure}[ht!]
  \centering
  \includegraphics[width=0.98\linewidth]{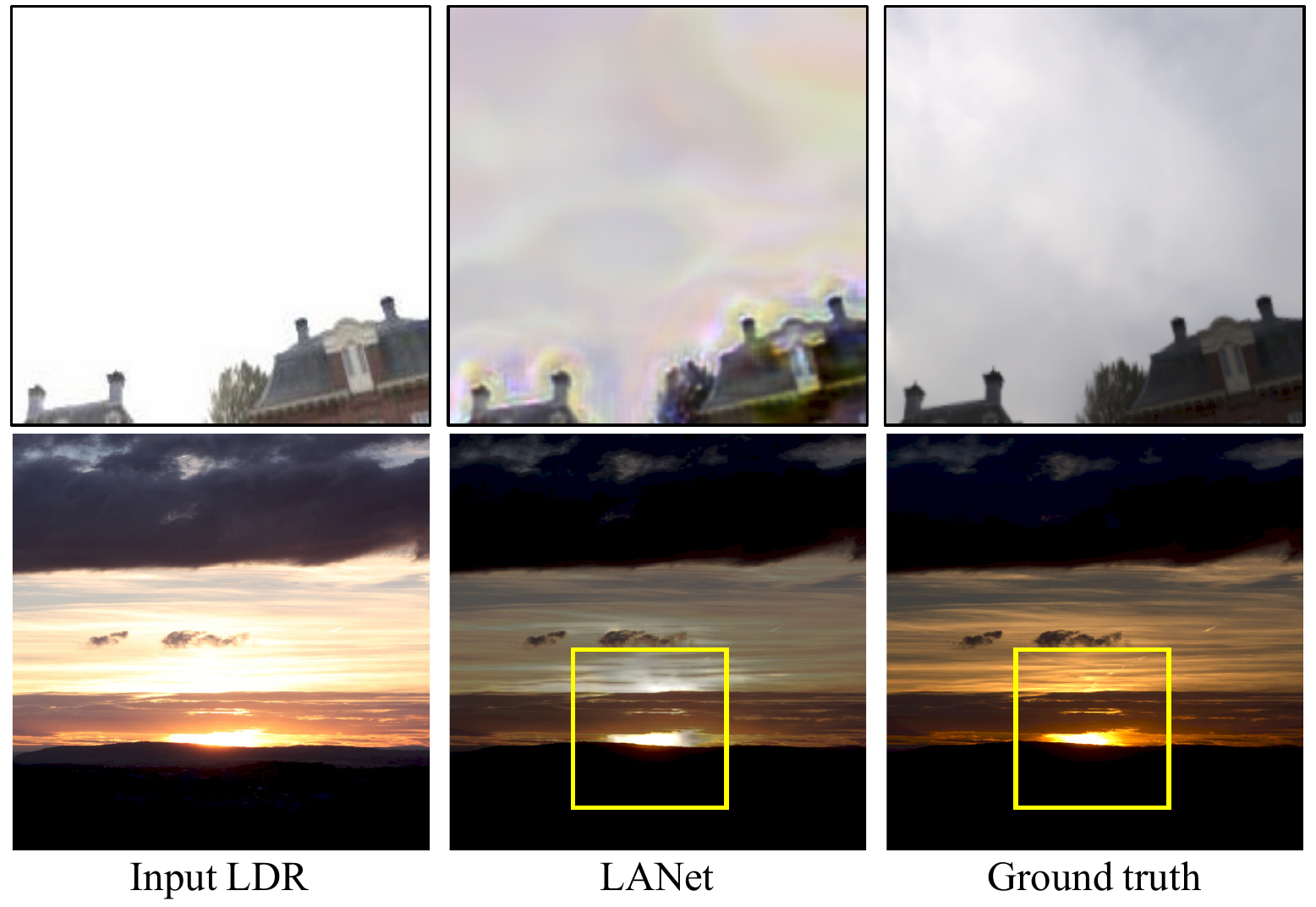}
  \caption{Failure cases. The top three images are with highly over-exposed areas that the textual detail cannot be recovered well. The bottom three images are without color information around the overexposed area, which leads to color information missing in the input LDR image for inferring HDR image.}
  \label{figure-fail}
  \vspace{-0.5cm}
\end{figure}

Although our method can reconstruct the high dynamic range well in many scenes, it still fails in some cases, especially in extremely over-exposure scenes. As shown in Fig.~\ref{figure-fail}, when there is a large over-exposure region in the image, our method cannot restore texture information and luminance information correctly. 

Regarding runtime, note that our training stage is carried out off-line, and here we only offer the runtime for the testing stage under our experimental environment. For per LDR image of the size $1920\times1080$ and the panorama of the size $1024\times 512$, it requires about 3 seconds and 1 second respectively to complete the testing throughout the trained model.

\section{Conclusion}
In this paper, we propose an end-to-end and trainable luminance attentive network for HDR image reconstruction from single LDR images, as well as its extended network for HDR panorama reconstruction for image-based lighting usage. Rather than using the general maximum normalization method on HDR data, we calibrate original HDR images to the similar luminance scale corresponding to LDR images. This treatment gives us many benefits on HDR reconstruction tasks, such as obtaining a set of HDR images with a similar luminance scale and getting pixel-level luminance segmentation labels automatically without requiring any extra manual annotation. Also, the designed luminance attention module can well explore the estimated luminance semantic segmentation to pay attention to over-exposure and under-exposure areas for better reconstructing the HDR images. The extended panoLANet also achieves a better performance on HDR panorama reconstruction for IBL usage.
Our future work includes combining deep learning-based HDR reconstruction with image inpainting to perform image restoration in the HDR domain and solve the texture loss of LDR images in over-exposed areas with no light source.

\section*{Acknowledgement}
\addcontentsline{toc}{section}{Acknowledgement}
This work was partly supported by the Key Technological Innovation Projects of Hubei Province (2018AAA062), NSFC (NO. 61972298, 61872277, 62171324), and CAAI-Huawei MindSpore Open Fund.

\input{LANet.bbl}

\bibliographystyle{eg-alpha-doi}

\newpage
\begin{appendices}

\section{Image Visualization}

\subsection{Visualizing HDR Images in LDR Format}
\begin{figure*}[h]
  \centering
  \includegraphics[width=0.99\textwidth]{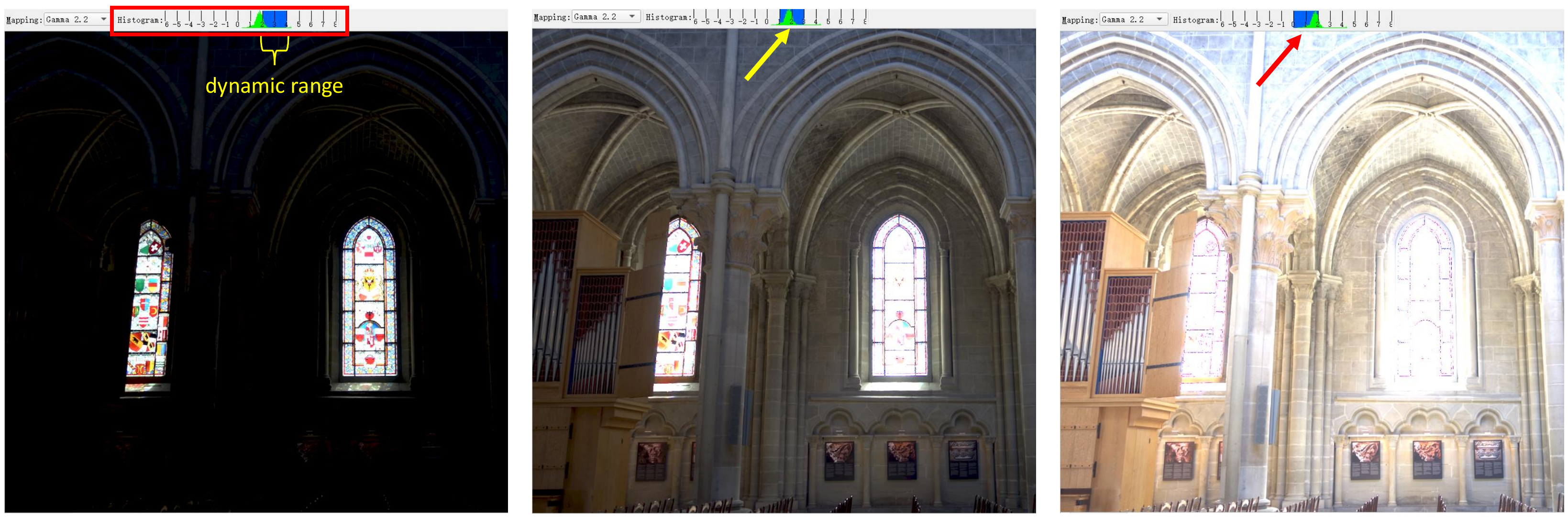}
  \caption{An HDR image shown in LDR format at different exposures. The histogram bar represents the distribution of the HDR image on the log scale. The range marked in blue indicates the dynamic mapping range and the position indicates the exposure. Note that here "Mapping: gamma 2.2" means mapping the linear result of LDR image on a display which gamma is $2.2$.}
  \label{hdr-visual}
\end{figure*}

As HDR images can't be displayed on common devices, we use linear mapping with a limited dynamic range at a certain exposure to compare HDR images in LDR format. Specifically, we use the preview window of the LuminanceHDR software ({\url{http://qtpfsgui.sourceforge.net}}) as the tool for visualizing HDR images. As shown in Fig.~\ref{hdr-visual}, we can use different exposures to present the overall information of HDR images as much as possible.

\subsection{Visualizing LDR Images at the Adjusted Exposure}
In order to highlight the difference between LDR images and HDR images more clearly, we also reduce or increase the exposure of LDR images for visualizing some figures in our paper. We complete the exposure reduction by scaling the LDR image in linear RGB space. Specifically, we first normalize the LDR image from the range $[0, 255]$ to $[0,1]$, and then convert it from sRGB color space to linear RGB color space using the following formula:
\begin{equation} \label{eq:sRGB2linear}
\begin{aligned}
I_{linear} &= \begin{cases}
 \frac{I_s}{12.92},& \text{ if } 0 \leq I_s \leq 0.04045 \\ 
 (\frac{(I_s + 0.055)}{1.055})^{2.4},& \text{ if } 0.04045 < I_s \leq 1
\end{cases}
\end{aligned} 
\end{equation}
Then we can adjust the exposure of LDR image in the same way as HDR image. For example, in Figure 1 of our paper, the left "Input LDR" shown on paper is the reduced version of the actual input image, and the right "Our result" is the output HDR image at the same dynamic range and reduced exposure.

\section{Datasets}
For the readers' convenience, we summarize the used datasets according to the data type in Table~\ref{tab:datasets}. We also provide the data source of each dataset.
\begin{table*}[ht!]
\normalsize
\centering
\caption{The list of HDR datasets we use in our experiments.}
\label{datalist}
\begin{tabular}{|c|l|l|c|}
    \hline
    Type & Dataset Name & Source & Number \\
    \hline
    \multirow{4}{*}{Pano} & Laval Indoor HDR Dataset & {\url{http://indoor.hdrdb.com}} & 2233 \\ 
    ~ & Laval Outdoor HDR Dataset & {\url{http://outdoor.hdrdb.com}} & 205 \\ 
    ~ & HDRI Haven & {\url{https://hdrihaven.com/hdris}} & 322 \\ 
    ~ & sIBL & {\url{http://www.hdrlabs.com/sibl/archive.html}} & 79 \\ 
    \hline
    \multirow{4}{*}{Img} & HDR Photographic Survey & {\url{http://rit-mcsl.org/fairchild/HDR.html}} & 105 \\ 
    ~ & Funt et al. HDR Dataset & {\url{https://www2.cs.sfu.ca/\~colour/data/funt\_hdr/\#DATA}} & 105 \\ 
    ~ & Stanford HDR Data & {\url{http://scarlet.stanford.edu/\~brian/hdr/hdr.html}} & 88 \\ 
    ~ & Ward & {\url{http://www.anyhere.com/gward/hdrenc/pages/originals.html}} & 33 \\ 
    ~ & HDR-Eye & {\url{https://www.epfl.ch/labs/mmspg/downloads/hdr-eye/}} & 42 \\ 
    \hline
    Video & LiU HDRv & {\url{http://hdrv.org/Resources.php}} & 10 \\ 
    \hline
\end{tabular}
\label{tab:datasets}
\end{table*}

\section{Network Architecture Details of LANet}
We implement our network architecture in Tensorflow and Tensorlayer. The resolution of the input images is any size greater than $256\times256$, and the output images are the same size as the input. During the training stage, the size of images is set to $256\times256$. Before inputting the images to the network, we first converted them from sRGB color space to linear RGB space using the above method, then normalized them to $[-1,1]$ as the final inputs for our network. The output HDR images from the network are in the logarithmic domain and need an exponential operation to get the final result. Here we describe our network architecture in detail. We first define some operations as follows:

\begin{itemize}
    \item[$\bullet$] $R_x$: Denoting the residual block in our network. We use the first five convolutional layers of ResNet50~\cite{he2016deep} with the version of "Relu before addition"~\cite{he2016identity} as the structure of five residual blocks in our network and define them in turn as $R_1$ to $R_5$.
    \item[$\bullet$] $C(s,k)$: Denoting Relu-Convolution-InstanceNorm layer with filters size $s \times s$ and output channels $k$. 
    \item[$\bullet$] $DC(s,k)$: Denoting Relu-Convolution-InstanceNorm layer with convolution stride $2$, filters size $s \times s$ and output channels $k$.
    \item[$\bullet$] $UC(s,k)$: Denoting Upsample-Relu-Convolution-InstanceNorm layer with a nearest-neighbor upsample which stride equals to $2$, filters size $s \times s$ and output channels $k$. 
    \item[$\bullet$] $SC_x(k)$: Denoting the skip connection layer with output channels set to $k$. For $SC_1$ to $SC_5$, the skip connections are used from $R_1$ to $R_5$ respectively. They first apply a $C(3,k)$ for each skip connection, then concatenate them with the output from last layer and apply a $C(1,k)$ to get the final outputs. For $SC_0$, it directly concatenates the network inputs with the LAM outputs, then applies a $C(3,k)$ to get the final network outputs.
\end{itemize}
Then the whole network with the HDR reconstruction stream is defined as:

(\textit{Inputs})\ \ -\ \ $R_1$\ \ -\ \ $R_2$\ \ -\ \ $R_3$\ \ -\ \ $R_4$\ \ -\ \ $R_5$\ \ -\ \ 

$DC(3,1024)$\ \ -\ \ $DC(3,1024)$\ \ -\ \ $UC(3,1024)$\ \ -\ \ 

$UC(3,512)$\ \ -\ \ $C(3,512)$\ \ -\ \ $SC_5(512)$\ \ -\ \ 

$UC(3,256)$\ \ -\ \ $C(3,256)$\ \ -\ \ $SC_4(256)$\ \ -\ \ 

$UC(3,128)$\ \ -\ \ $C(3,128)$\ \ -\ \ $SC_3(128)$\ \ -\ \ 

$UC(3,64)$\ \ -\ \ $C(3,64)$\ \ -\ \ $SC_2(64)$\ \ -\ \ 

$UC(3,64)$\ \ -\ \ $C(3,64)$\ \ -\ \ $SC_1(64)$\ \ -\ \ 

LAM\ \ -\ \ $SC_0(3)$\ \ -\ \ (\textit{Outputs})

\noindent And the luminance segmentation stream is defined as:

($SC_3(128)$)\,\,\,-\,\,\,$UC(3,64)$\,\,\,-\,\,\,$C(3,64)$\,\,\,-\,\,\,$UC(3,64)$\,\,\,-\,\,\,$UC(3,3)$\,\,\,-\,\,\,(\textit{Seg. Outputs})

\section{More Visual Comparision Results}
We show more qualitative comparison detail results of predicted HDR images in Figure 14 to 17 and results of predicted HDR panoramas in Figure 18 to 20, which represent the performance of our method under different exposure conditions.



\clearpage
\begin{figure*}[h]
  \centering
  \includegraphics[width=0.86\textwidth]{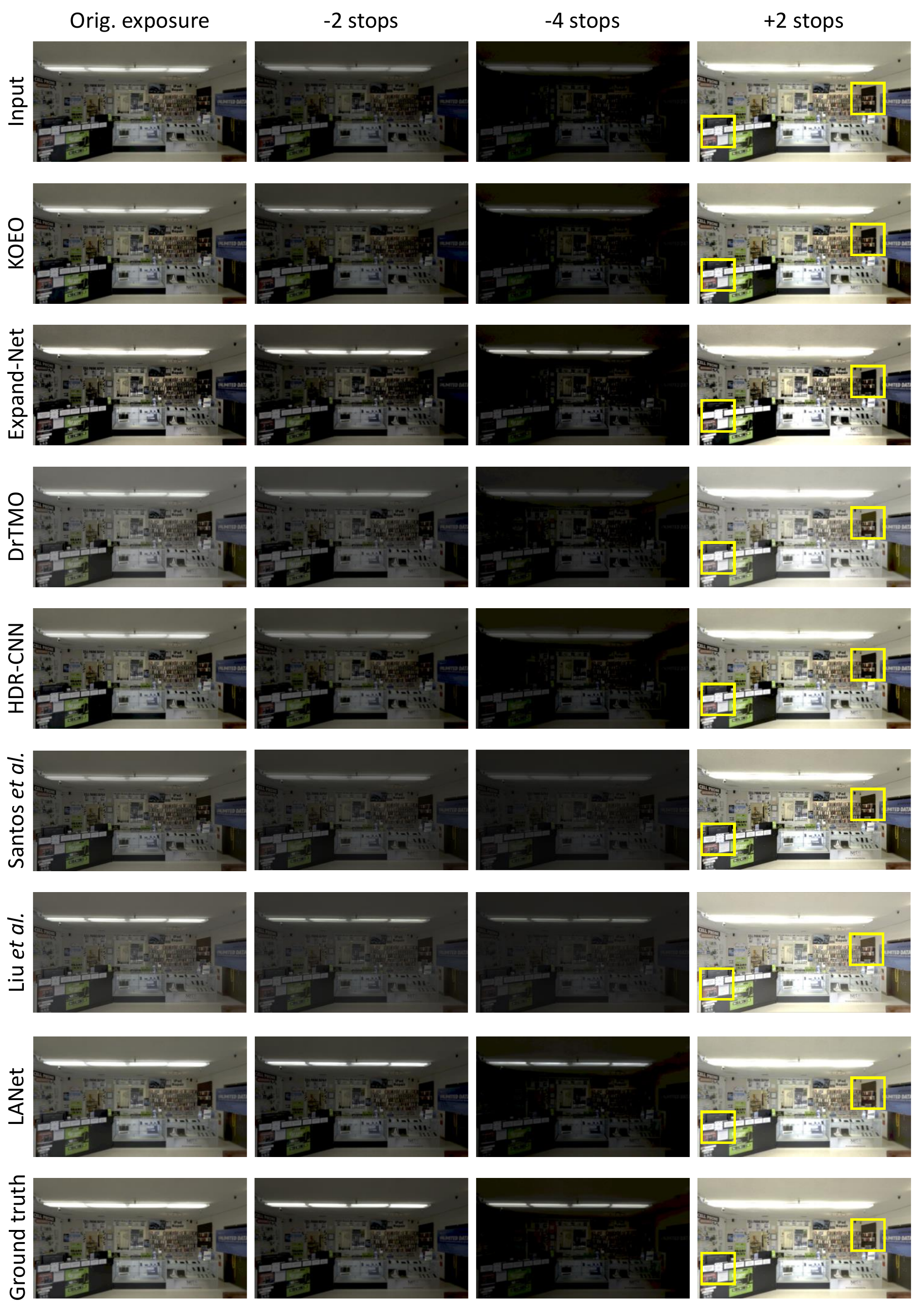}
  \caption{Comparision on indoor scene at different visualization exposures.
  \label{compare1}}
\end{figure*}

\clearpage
\begin{figure*}[h]
  \centering
  \includegraphics[width=0.86\textwidth]{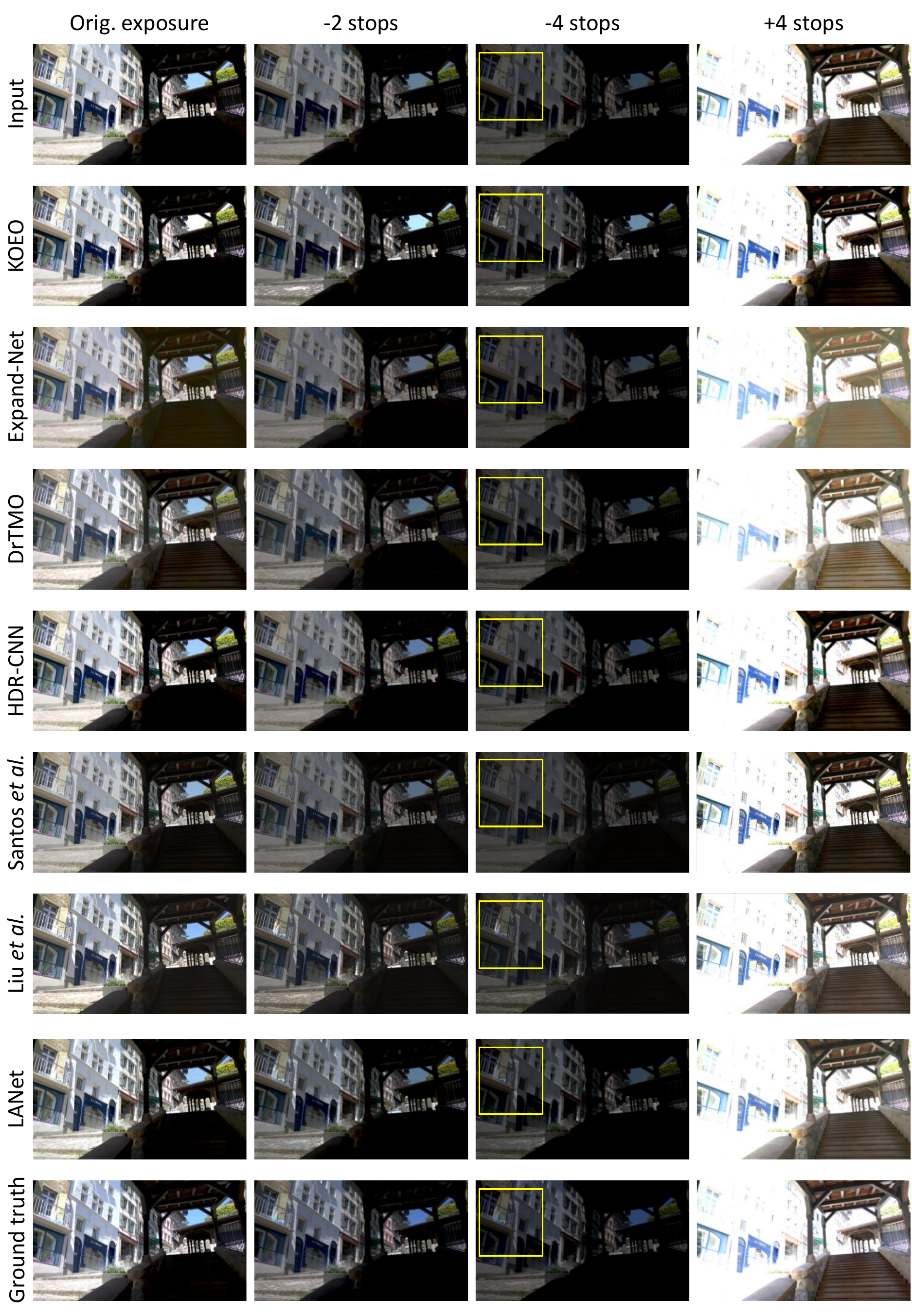}
  \caption{Comparision on outdoor scene at different visualization exposures.}
  \label{compare2}
\end{figure*}

\clearpage
\begin{figure*}[h]
  \centering
  \includegraphics[width=0.86\textwidth]{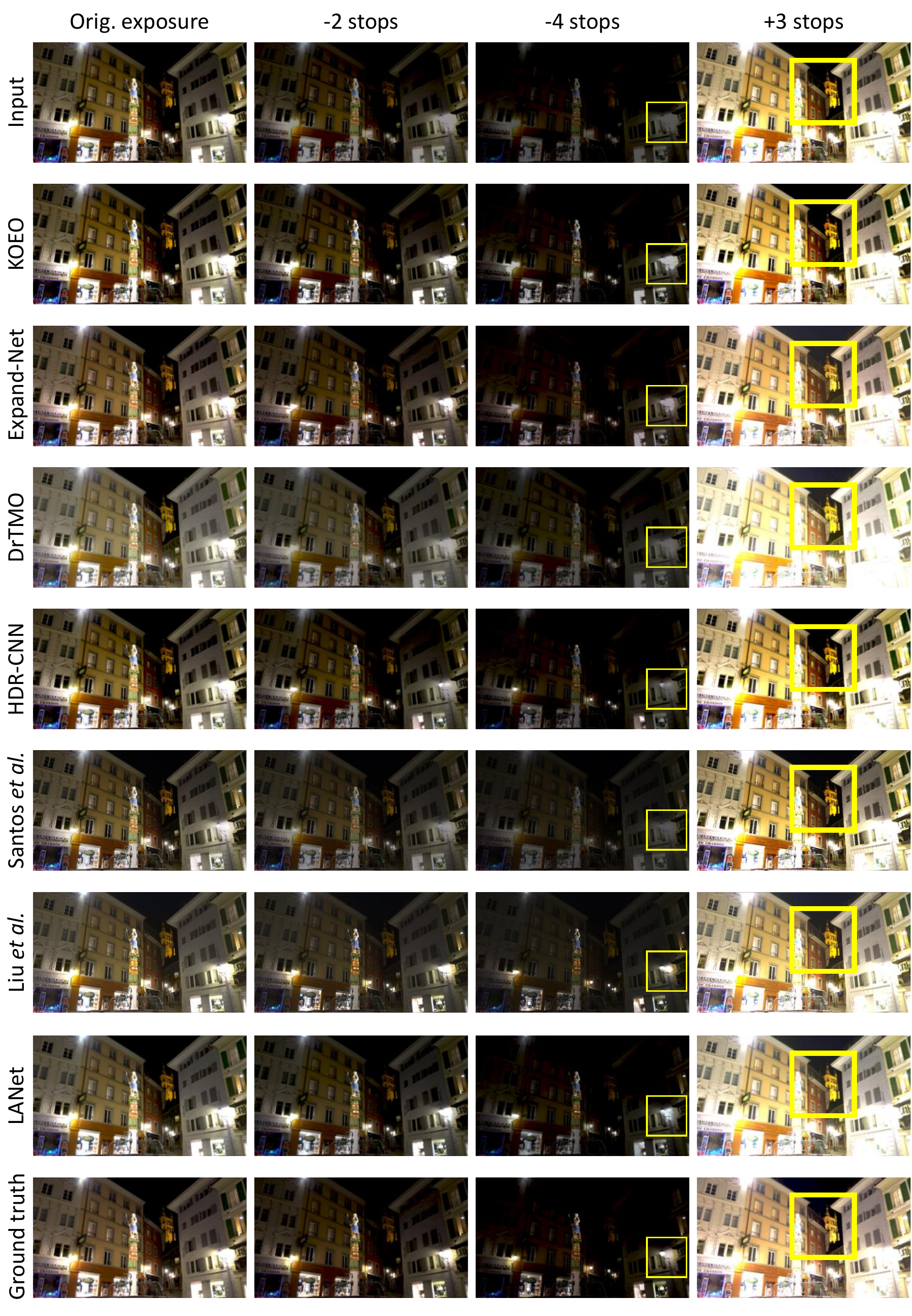}
  \caption{Comparision on night scene at different visualization exposures.}
  \label{compare3}
\end{figure*}

\clearpage
\begin{figure*}[h]
  \centering
  \includegraphics[width=0.86\textwidth]{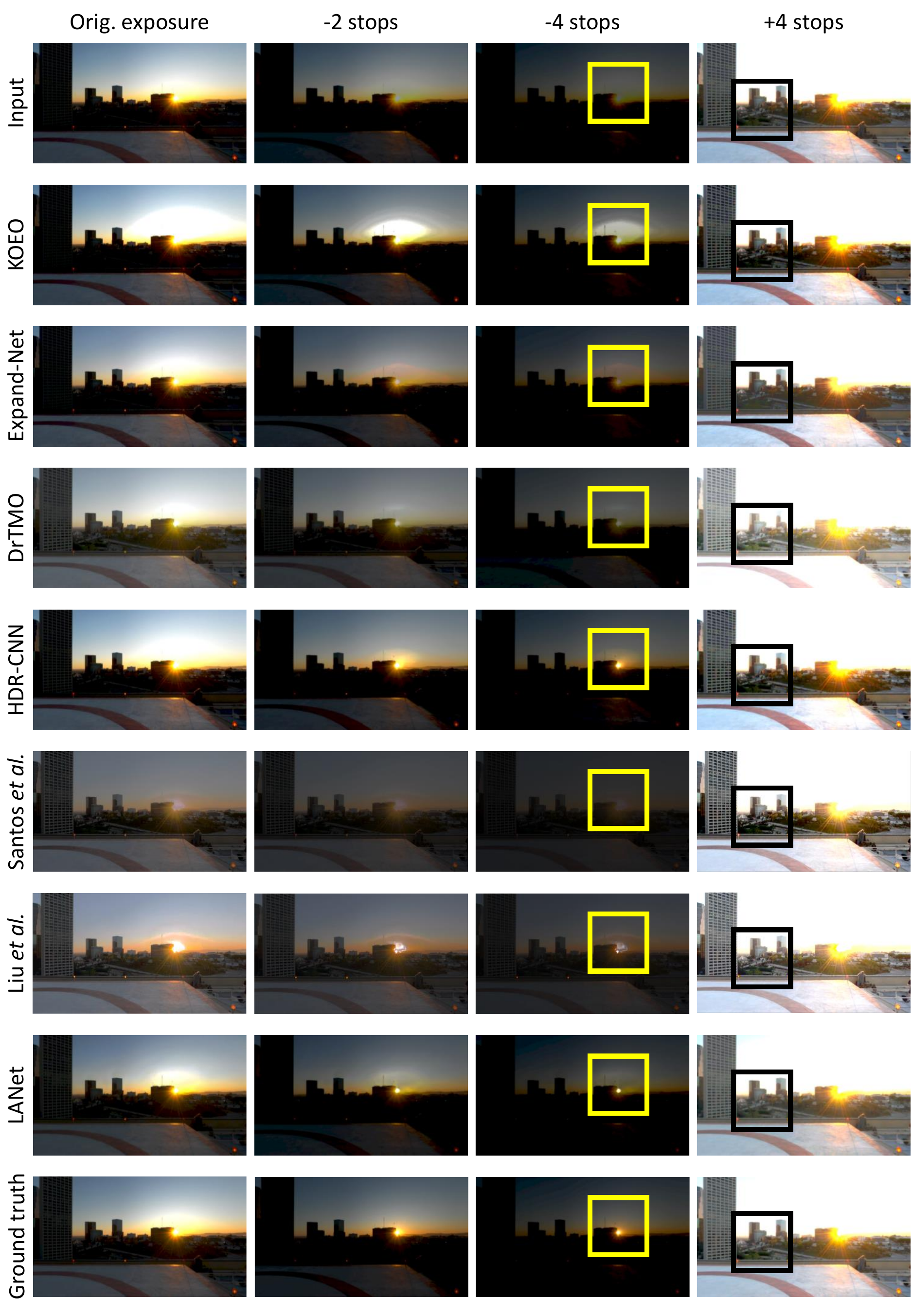}
  \caption{Comparision on extreme highlight scene at different visualization exposures.}
  \label{compare4}
\end{figure*}

\clearpage
\begin{figure*}[h]
  \centering
  \includegraphics[width=0.84\textwidth]{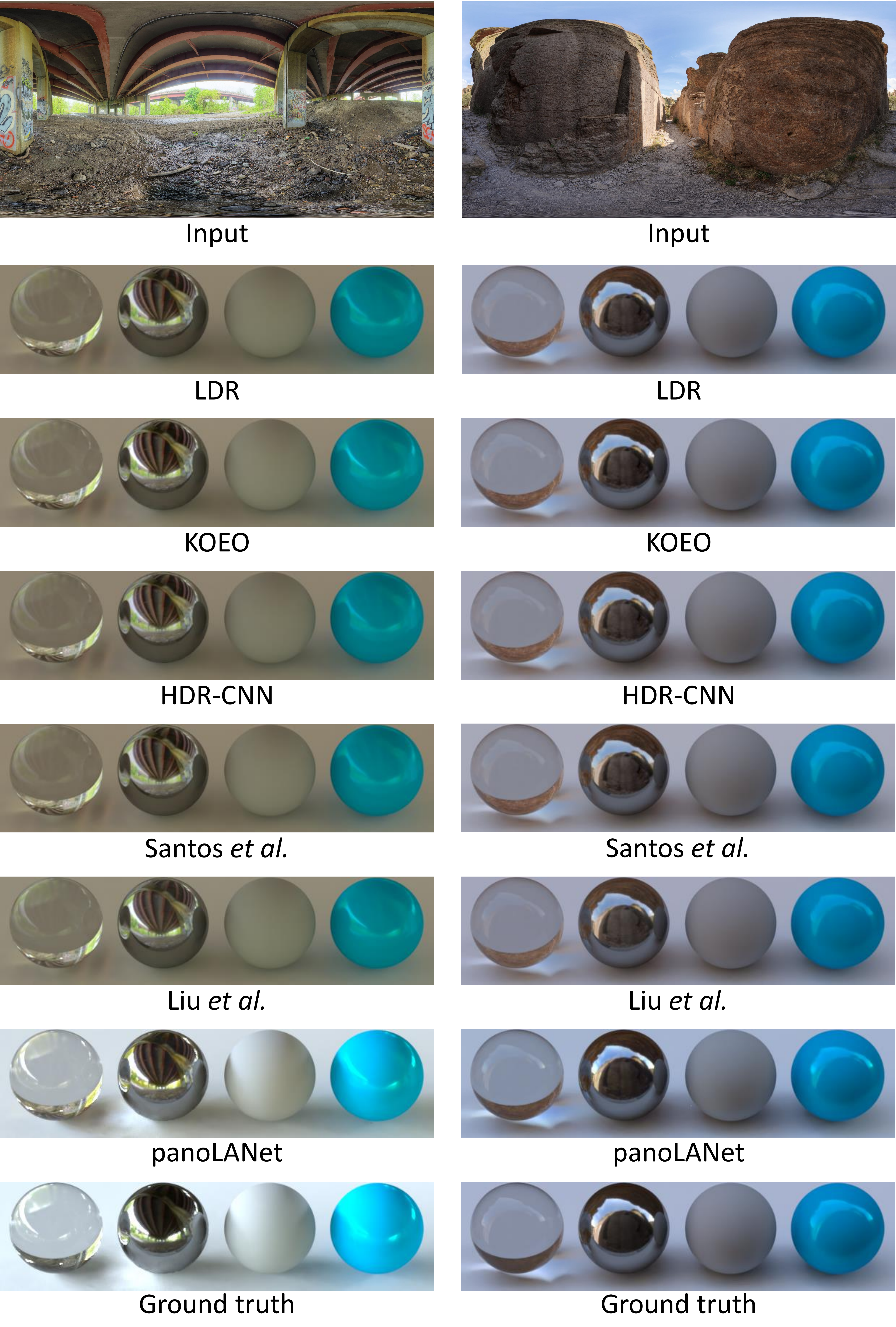}
  \caption{Comparision on rendering result of predicted panoramas.}
  \label{compare5}
\end{figure*}

\clearpage
\begin{figure*}[h]
  \centering
  \includegraphics[width=0.84\textwidth]{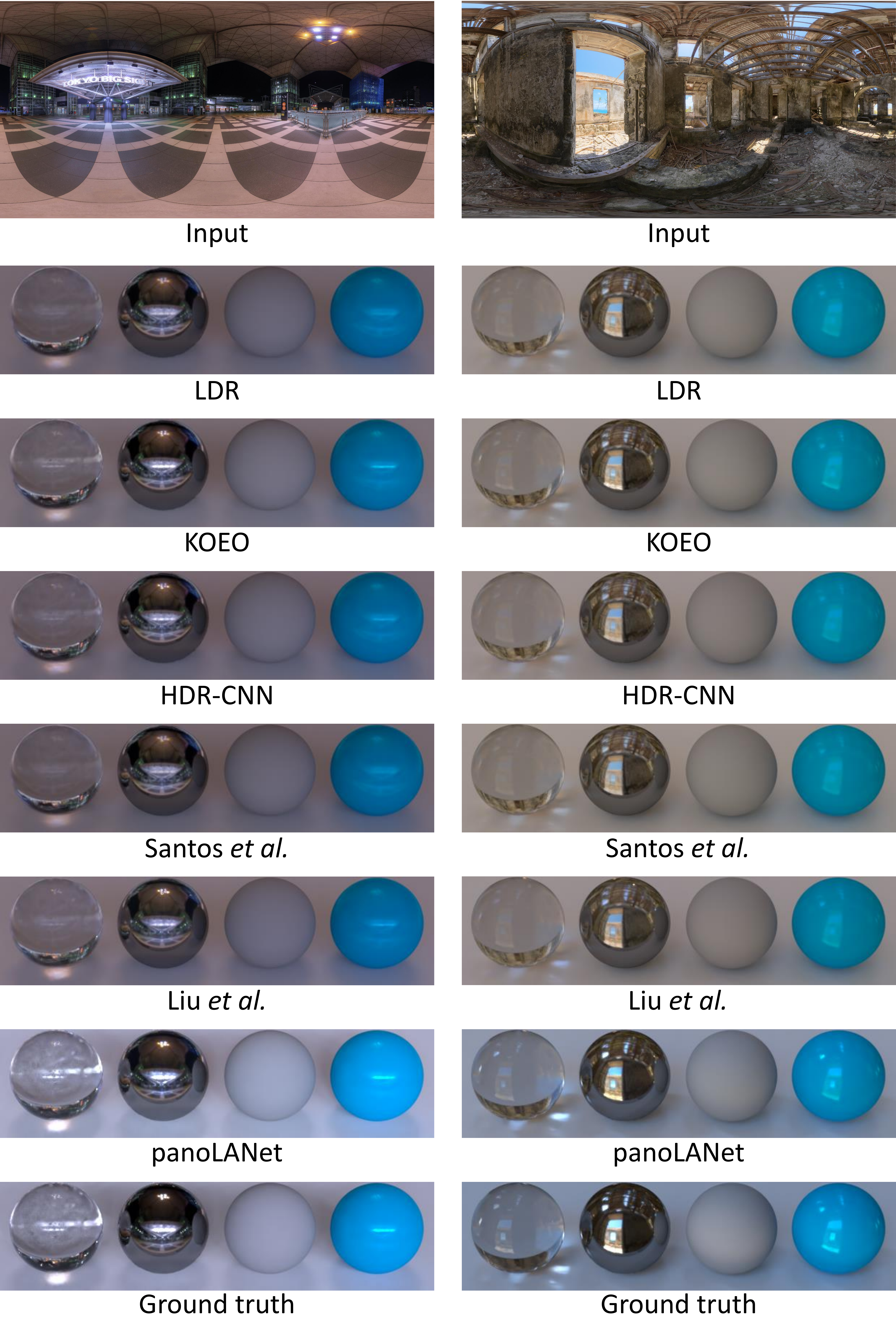}
  \caption{Comparision on rendering result of predicted panoramas.}
  \label{compare6}
\end{figure*}

\clearpage
\begin{figure*}[h]
  \centering
  \includegraphics[width=0.84\textwidth]{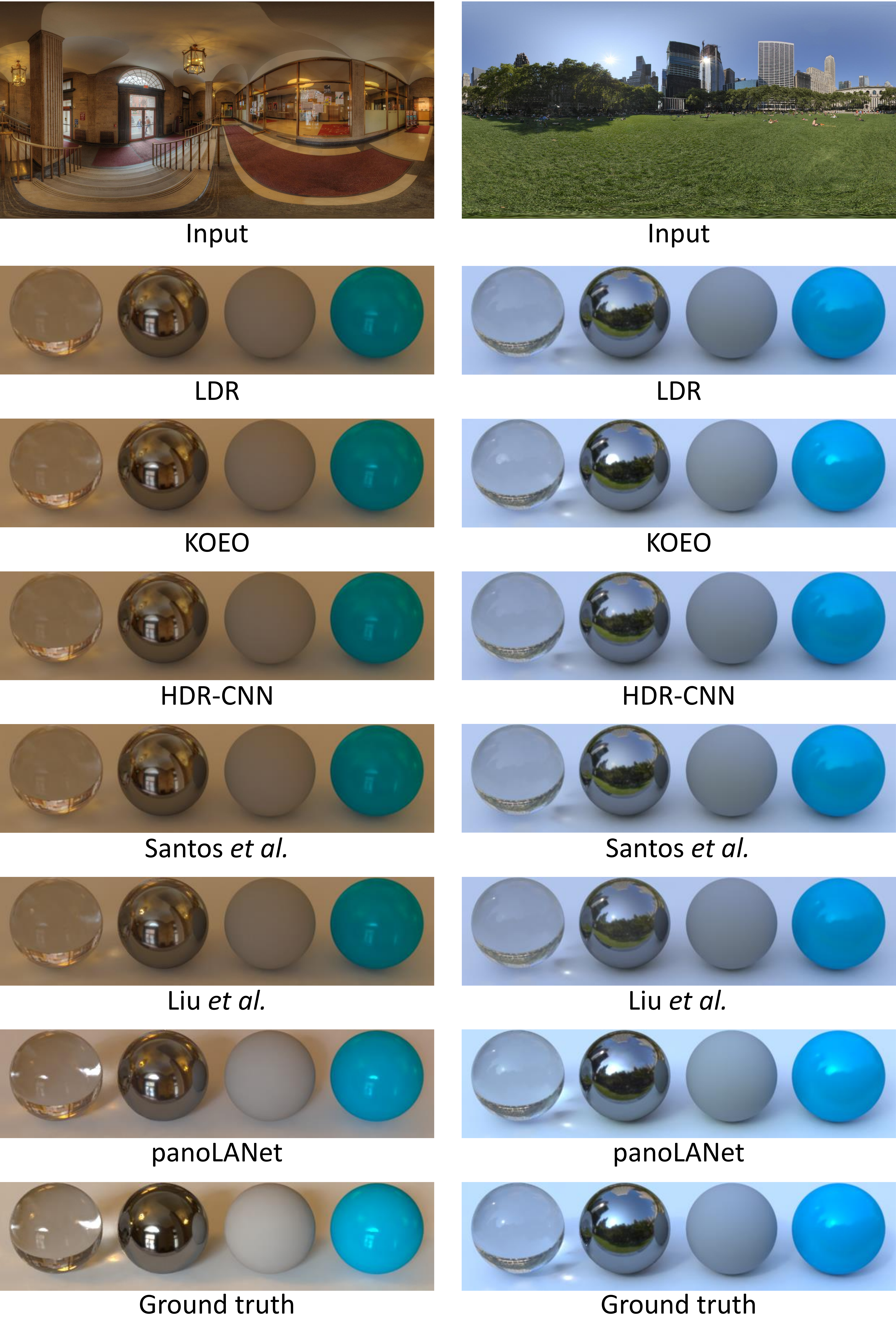}
  \caption{Comparision on rendering result of predicted panoramas.}
  \label{compare7}
\end{figure*}

\end{appendices}

\end{document}

%% file: LANet.bbl
\newcommand{\etalchar}[1]{$^{#1}$}